\documentclass[twoside]{article}
\pdfoutput=1
\usepackage[utf8]{inputenc} 
\usepackage[T1]{fontenc}    
\usepackage{hyperref}       
\usepackage{url}            
\usepackage{booktabs}       
\usepackage{amsfonts}       
\usepackage{nicefrac}       
\usepackage{microtype}      
\usepackage{xcolor}         

\usepackage{aligned-overset}
\usepackage{bbm}
\usepackage{bm}
\usepackage{calrsfs}
\usepackage{colonequals}
\usepackage{centernot}
\usepackage{xfrac}
\usepackage{url}

\usepackage{amsmath}
\usepackage{amssymb}
\usepackage{mathtools}
\usepackage{amsthm}

\usepackage{microtype}
\usepackage{graphicx}
\usepackage[subrefformat=parens]{subcaption}
\usepackage{wrapfig}
\usepackage[rightcaption]{sidecap}
\usepackage{multirow}
\usepackage[inline]{enumitem}
\usepackage{colortbl}

\usepackage{pgfplots}
\DeclareUnicodeCharacter{2212}{−}
\usepgfplotslibrary{groupplots,dateplot}
\usetikzlibrary{patterns,shapes.arrows}
\pgfplotsset{compat=newest}

\usepackage{todonotes}

\usepackage[accepted]{aistats2024}
\usepackage[round]{natbib}

\theoremstyle{plain}
\newtheorem{theorem}{Theorem}[section]
\newtheorem{proposition}[theorem]{Proposition}

\theoremstyle{definition}

\theoremstyle{remark}

\newcommand{\expectation}{\mathbb{E}}

\newcommand{\simplex}[1]{\Delta^{#1}}

\definecolor{graygray}{RGB}{229,229,229}

\begin{document}

%
\runningtitle{Consistent and Asymptotically Unbiased Estimation of Proper Calibration Errors}

%
\runningauthor{Popordanoska, Gruber, Tiulpin, Buettner, Blaschko}

\twocolumn[

\aistatstitle{Consistent and Asymptotically Unbiased Estimation of Proper Calibration Errors}

\aistatsauthor{Teodora Popordanoska$^{*}$ \And Sebastian G. Gruber$^{*}$}
\aistatsaddress{ESAT-PSI \\ KU Leuven, Belgium \And German Cancer Research Center (DKFZ) \\
German Cancer Consortium (DKTK) \\
Goethe University Frankfurt, Germany }

\aistatsauthor{Aleksei Tiulpin \And Florian Buettner \And Matthew B. Blaschko}
\aistatsaddress{ HST Research Unit \\ Faculty of Medicine \\University of Oulu, Finland \And German Cancer Research Center (DKFZ) \\
German Cancer Consortium (DKTK) \\
Frankfurt Cancer Institute, Germany \\
Goethe University Frankfurt, Germany \And ESAT-PSI \\ KU Leuven, Belgium} 
]

\begin{abstract}
Proper scoring rules evaluate the quality of probabilistic predictions, playing an essential role in the pursuit of accurate and well-calibrated models. 
Every proper score decomposes into two fundamental components -- \textit{proper calibration error} and refinement -- utilizing a Bregman divergence.
While uncertainty calibration has gained significant attention, current literature lacks a general estimator for these quantities with known statistical properties. 
To address this gap, we propose a method that allows consistent, and asymptotically unbiased estimation of \textit{all} proper calibration errors and refinement terms. 
In particular, we introduce Kullback--Leibler calibration error, induced by the commonly used cross-entropy loss. 
As part of our results, we prove the relation between refinement and f-divergences, which implies information monotonicity in neural networks, regardless of which proper scoring rule is optimized.
Our experiments validate empirically the claimed properties of the proposed estimator and suggest that the selection of a post-hoc calibration method should be determined by the particular calibration error of interest.
\end{abstract}

\section{Introduction}

Risk minimization is the cornerstone of machine learning, where the goal is to develop models that are accurate and provide well-calibrated uncertainty estimates.
Central to this pursuit are proper scoring rules \citep{gneitingscores}, which measure the quality of probabilistic predictions via dissimilarity measures of probability distributions, known as Bregman divergences \citep{bregman1967relaxation,ovcharov2018proper}.
A significant breakthrough in this realm is the decomposition of the expected loss associated with a proper score into calibration and refinement \citep{ANewVectorPartitionoftheProbabilityScore, degroot1981assessing,blattenberger1985separating,Br_cker_2009}. Facilitated by this result, understanding and mitigating calibration error (CE) has become one of the key concerns for applications like healthcare~\citep{HAGGENMULLER2021202,KatsaouniTashkandiWieseSchulz+2021+871+885}, climate modelling~\citep{Gneiting2005WeatherFW,kashinath2021physics} and autonomous driving~\citep{yurtsever2020survey}. 

On the most fundamental level, calibration errors compare a predictive distribution with a conditional target distribution \citep{gruber2022better}. For this, the machine learning literature proposed a wide range of different calibration errors in the multi-class setting \citep{Br_cker_2009, kull2015novel, naeini2015, vaicenavicius2019evaluating, kumar2019verified, zhang2020mix, popordanoska2022, gruber2022better}. The most common ones are based on absolute or squared differences. However, current literature lacks a general estimator of calibration error and refinement induced by \textit{any} Bregman divergence. 

\begin{table*}[ht!]
    \caption{
    Proposed estimators of calibration error and refinement, induced by Brier score (first row) and log loss (second row) for a classifier $g$, one-hot encoded label $y$, and dataset size $n$. The term $\widehat{\mathbb{E}[Y \mid g(x_h)]}$ is defined in Equation \eqref{eq:CondExpectRatioEstimateKDE} via kernel density estimation.
    }
    \label{tab:formula_summary_table1}
    \centering
    \resizebox{0.90\textwidth}{!}{
    \begin{tabular}{cccc}
        \toprule
         Loss $L\left( g \left( x \right), y \right)$
         & Calibration error estimator $\widehat{\operatorname{CE}}_F(g)$ & Refinement estimator $\widehat{\operatorname{REF}}_F(g)$ \\
        \midrule
         $\left\lVert g \left( x \right) - y \right\rVert_2^2$ &
         $\frac{1}{n} \sum\limits_{h=1}^n  \left\| \widehat{\mathbb{E}[Y \mid g(x_h)]} - g(x_h) \right\|_2^2 $ & 
         $-\frac{1}{n} \sum\limits_{h=1}^n  \left\| \widehat{\mathbb{E}[Y \mid g(x_h)]} \right\|_2^2 $\\
        \addlinespace  
        \addlinespace  
         $- \left\langle \log g \left( x \right), y \right\rangle$ &
         $\frac{1}{n} \sum\limits_{h=1}^n \left\langle  \widehat{\mathbb{E}[Y \mid g(x_h)]}, \log \frac{\widehat{\mathbb{E}[Y \mid g(x_h)]}}{g(x_h)} \right\rangle$ &
        $-\frac{1}{n}\sum\limits_{h=1}^n 
        \left\langle  \widehat{\mathbb{E}[Y \mid g(x_h)]}, \log \widehat{\mathbb{E}[Y \mid g(x_h)]} \right\rangle $ \\
        \bottomrule
    \end{tabular}
    }
\end{table*}   

We approach the study of calibration errors through the notion of \textbf{proper calibration errors} \citep{gruber2022better}, which is a general class of calibration errors derived from risk minimization via proper scores.
For example, the Brier score induces the squared $L_2$ calibration error, for which there exists a consistent estimator \citep{popordanoska2022}.
Estimating the KL calibration error, which is induced by the most common proper score in classification -- the categorical negative log likelihood\footnote{We use the terms categorical negative log-likelihood, log-loss and cross-entropy loss interchangeably.}  remains an open challenge.

In this work, we introduce a \textit{consistent and asymptotically unbiased} estimator for all proper calibration errors and refinement terms. 
Additionally, our proposed estimator can also be used for estimating the so-called model sharpness. Similar to how every proper score generates a proper calibration error, a sharpness term is generated.
For example, the sharpness induced by the log loss is the mutual information between prediction and target variable \citep{huszar2013scoring}.
It is investigated in the context of general forecasts, but is not well understood for neural networks \citep{morris1983comparison, allan1977reliability, blattenberger1985separating, Br_cker_2009, ANewVectorPartitionoftheProbabilityScore}.
We show that any model sharpness is identical to an f-divergence \citep{csiszar1972class}.
Through the information monotonicity of the f-divergence, we then derive the concept of general information monotonicity in neural networks.
This gives a novel perspective into the workings of neural networks and illustrates how the information bottleneck theory is a more general concept beyond mutual information.
The source codes will be released upon acceptance.

Our \textbf{contributions} can be summarized as follows:
\begin{enumerate}[noitemsep]
    \item 
    We provide a general estimator for all proper calibration errors and refinement in classification, which is consistent and its bias converges with rate $\mathcal{O} \left( n^{-1} \right)$. In Table~\ref{tab:formula_summary_table1} we show our derived estimators for calibration error and refinement induced by Brier score and log loss.
    \item 
    We show that model sharpness can be formulated as a multi-distribution f-divergence and is upper bounded by the statistical information of the classification task. Based on this result, we derive the concept of information monotonicity in neural networks beyond mutual information.
    \item 
    We conduct experiments showcasing the empirical properties of our estimator, as well as its utility in selecting an appropriate post-hoc calibration method for the desired calibration error.
\end{enumerate}

\section{Related work}
\label{sec:background}

In this section, we give a brief overview of estimating calibration errors.
Calibration errors are notoriously difficult to estimate since they compare a prediction $g \left( X \right)$ with the conditional expectation $\mathbb{E} \left[ Y \mid g \left( X \right) \right]$, where $Y$ is the one-hot encoded target variable, $X$ the input variable, and $g$ the predictive model.
The difficulty arises since $g \left( X \right)$ is in general a multivariate continuous random variable.
Originally, model calibration was only considered for a finite set of predictions \citep{ANewVectorPartitionoftheProbabilityScore}, simplifying the estimation since $g \left( X \right)$ is then a discrete random variable.
\citet{Platt99probabilisticoutputs} used histogram estimation, which transforms the continuous prediction space to a discrete one, to assess the calibration of a continuous binary model.
Different ways to define the histogram bins are equal width or equal mass binning techniques \citep{nguyen2015posterior}.
The number of bins and the binning scheme can significantly influence the estimated value \citep{kumar2019verified} and there is no optimal default since every setting has a different bias-variance tradeoff \citep{nixon2019measuring}.
Further, using a fixed binning scheme represents a lower bound of the respective calibration error \citep{kumar2019verified, vaicenavicius2019evaluating}.
In consequence, \cite{vaicenavicius2019evaluating} propose to use adaptive binning similar to other approaches in histogram estimation \citep{nobel1996histogram}.
\citet{roelofs2022mitigating} also introduce an algorithm to optimize the number of bins.
\citet{zhang2020mix} circumvent binning schemes by using kernel density estimation via Bayes' theorem. \\
The first calibration estimator for a multi-class model was given by \citet{naeini2015} and is still the most commonly used measure to quantify calibration, known as expected calibration error (ECE) \citep{guo2017calibration}. \\
The ECE is a special case of top-label confidence calibration since it only uses the predicted top-label confidence $\max_{i \in \mathcal{Y}} g_i \left( X \right)$ and compares it with the conditional accuracy $\mathbb{E} \left[ Y_C \mid \max_{i \in \mathcal{Y}} g_i \left( X \right) \right]$ with $C = \arg\max_{i \in \mathcal{Y}} g_i \left( X \right)$.
In contrast, class-wise calibration estimation uses all predicted classes, but only in isolation from each other since it compares $g_i \left( X \right)$ with $\mathbb{E} \left[ Y_i \mid g_i \left( X \right) \right]$ for each class $i$.
Both notions depend on estimating a conditional distribution given a univariate continuous random variable.
Consequently, estimation schemes of one notion can be applied to the other.
For example, \citet{kull2019beyond} and \citet{nixon2019measuring} use histogram binning estimation to estimate class-wise calibration. In contrast, canonical calibration refers to the case when the model prediction $g \left( X \right)$ matches the conditional target $\mathbb{E} \left[ Y \mid g \left( X \right) \right]$ almost surely \citep{vaicenavicius2019evaluating, popordanoska2022}.
It is substantially more difficult to estimate with increasing classes since $g \left( X \right)$ also increases in dimensionality.
This makes histogram based approaches infeasible and kernel density estimation strongly dependent on the scalability of the kernel to higher dimensions.
\citet{popordanoska2022} propose a specific kernel choice to estimate canonical $L_p$ calibration errors.

\section{Proper calibration errors}

In classification, it is common to use a loss function of the form $L \colon \Delta^k \times \mathcal{Y}$, where $\Delta^k$ is the $(k-1)$ dimensional simplex and $\mathcal{Y}$ the sample space of the one-hot encoded target variable $Y$.
Further, assume $X$ is the feature variable with realizations in a space $\mathcal{X}$.
To optimize a model $g \colon \mathcal{X} \to \Delta^k$ mapping from the feature space into the probability simplex, we use the expected loss
\begin{equation}
   \mathcal{R} \left( g \right) \coloneqq \mathbb{E} \left[ L \left( g \left( X \right), Y \right) \right],
\end{equation}
which is referred to as \textbf{risk}.
If $L$ is minimized by the Bayes classifier $g^* \left( x \right) \coloneqq \mathbb{E} \left[ Y \mid X = x \right]$, then $-L$ is a proper score \citep{gneitingscores}.
In that case, we may also refer to $L$ as a proper loss \citep{williamson2014geometry}.
The best achievable (negative) risk for a given target distribution $q$ is given by $F \left( q \right) \coloneqq - \inf_{p \in \Delta^k} \mathbb{E}_{Y \sim q} \left[ L \left( p, Y \right) \right]$.
The function $F$ is convex if and only if $-L$ is a proper score \citep{gneitingscores}.

Every differentiable proper score is uniquely associated with a Bregman divergence.
A Bregman divergence \citep{bregman1967relaxation} in a $k$-dimensional space $U \subset \mathbb{R}^k$ is characterized by a continuously differentiable, strictly convex function $F \colon U \to \mathbb{R}$, with
$D_F (p,q) \coloneqq F(p) - F(q) - \left\langle \nabla F(q), p-q \right\rangle$.
Special cases are the squared Euclidean distance with $F(p) = \| p \|_2^2$, and the Kullback--Leibler divergence with $F(p) = \sum_{i=1}^k p_i \log \left( p_i \right)$.  

Following \citet{ovcharov2018proper}, the risk related to a proper loss is connected to a Bregman divergence via
\vskip -0.1in
\begin{equation}
    \mathcal{R} \left( g \right) + \mathbb{E} \left[ F \left( \mathbb{E} \left[ Y \mid X \right] \right) \right] = \mathbb{E} \left[ D_F \left( \mathbb{E} \left[ Y \mid X \right], g \left( X \right) \right) \right].
\end{equation}

Consequently, risk minimization is equivalent to minimizing an expected Bregman divergence since they only differ by a model independent constant.
The most prominent example is the log loss, which induces the Kullback--Leibler divergence via the Shannon entropy \citep{ovcharov2018proper}. 

We can use Bregman divergences to assess the canonical calibration of a model.
\citet{Br_cker_2009} showed that proper scores in classification can be decomposed into calibration and refinement terms.
Similarly, we can do the same for the risk, namely
\begin{align}
\label{eq:RiskCESharp}
    \mathcal{R} \left( g \right) = & \underbrace{\mathbb{E} \left[ D_F \left( \mathbb{E} \left[ Y \mid g \left( X \right) \right], g \left( X \right) \right) \right]}_{\text{Calibration}} \notag \\
    & + \underbrace{\mathbb{E} \left[ - F \left( \mathbb{E} \left[ Y \mid g \left( X \right) \right] \right) \right]}_{=: \operatorname{REF}_F \left( g \right) \text{ (Refinement)}}.
\end{align}

\citet{gruber2022better} defined a calibration error of the form 
\begin{equation}
\operatorname{CE}_F \left( g \right) \coloneqq \mathbb{E} \left[ D_F \left( \mathbb{E} \left[ Y \mid g \left( X \right) \right], g \left( X \right) \right) \right]    
\end{equation}
as \textbf{proper calibration error}.
Since any convex function defined on the simplex can be related to a proper score \citep{ovcharov2015existence}, any Bregman divergence can be used to define a proper calibration error.
\citet{ANewVectorPartitionoftheProbabilityScore} derived the squared calibration error from the Brier score.
It is defined as
\begin{align}\label{eq:L2calibrationError}
    \operatorname{CE}_2^2 (g) = \mathbb{E}\biggl[\Bigl\| \mathbb{E}[Y \mid g(X)]-g(X)\Bigr\|_2^2\biggr].
\end{align}

With the square root applied, it is also member of the so-called $L_p$ calibration errors, where the $2$-norm is replaced by a general $p$-norm \citep{naeini2015,kumar2019verified,wenger2020,popordanoska2022}.
\citet{popordanoska2022} propose an estimator of $L_p$ calibration errors via kernel density estimation and offer evaluations of the squared calibration error.
Another example of a proper calibration error can be derived from the log-likelihood and results in a calibration error based on the Kullback--Leibler divergence $D_{\mathrm{KL}}$ given by
\vskip -0.1in
\begin{equation}
    \operatorname{CE}_{\mathrm{KL}} (g) = \mathbb{E}\left[ D_{\mathrm{KL}} \left( \mathbb{E}[Y \mid g(X)], g(X) \right) \right].
\end{equation}

As all differentiable proper calibration errors are induced by a Bregman divergence, non-negativity applies immediately, but the range is dependent on which divergence is employed: $\operatorname{CE}_2^2(g) \in [0,2]$, while $\operatorname{CE}_{\mathrm{KL}} (g) \in [0,\infty )$.
Since we are only using distributions as inputs for calibration errors, the Kullback--Leibler CE is more principled according to information theory than the squared calibration error \citep{mackay2003information}.
The squared error implies a Euclidean geometry for its inputs since the input space is non-bounded.
But, distributions are non-negative and normalized, and, consequently, exist in a bounded space.
The Kullback--Leibler divergence is a better representation of these restrictions since it is also not defined for negative inputs \citep{mackay2003information}.
Currently, no estimator for general proper calibration errors exist, including the Kullback--Leibler case.
As part of this work we propose a consistent, asymptotically unbiased and differentiable estimator for any proper calibration error.

\section{Estimating proper calibration errors}
\label{sec:estimating_proper_ce}

Given an i.i.d.\ labeled data sample $\{(x_i,y_i)\}_{1\leq i\leq n}$, a generic Bregman divergence calibration error estimator can be defined via
\begin{align}
\label{eq:BregmanCalibrationError}
    \operatorname{CE}_F(g) 
    \approx  \frac{1}{n} \sum_{h=1}^n \Bigg( & F\left( \widehat{\mathbb{E}[Y \mid g(x_h)]} \right) - F(g(x_h))  \\ \nonumber 
    -& \left\langle \nabla F \left(g(x_h) \right) , \widehat{\mathbb{E}[Y \mid g(x_h)]}  - g(x_h)  \right\rangle \Bigg).
\end{align}
It is straightforward to verify that the application of $F(x) = \|x\|_2^2$ recovers exactly the $L_2$ special case of the estimator given by \cite[Equation~(9)]{popordanoska2022}, while setting $F(p) = \langle p, \log(p)\rangle$ yields
\begin{align}
\operatorname{CE}_{\mathrm{KL}}(g) \approx& 
\frac{1}{n} \sum_{h=1}^n    \left\langle \widehat{\mathbb{E}[Y \mid g(x_h)]}, \log \left( \frac{\widehat{\mathbb{E}[Y \mid g(x_h)]}}{g(x_h)}\right) \right\rangle, \label{eq:KLcalibrationError} 
\end{align}
where $\log$ and division operations are taken element-wise, and we may interpret the inner product using $\lim_{y\searrow 0} y \log y = 0$ to ensure that it remains well defined on the vertices of the probability simplex. Detailed derivations of calibration error and refinement estimators are in Appendix~\ref{appendix:bregman_divergence}. In the sequel, our notation will use capital letters for unbounded random variables, and lower case variables with subscripts for elements of our data sample. Note that we may still treat elements of the data sample as random variables.

We define the finite sample estimator for the conditional expectation as in \citet[Equation~(3)]{popordanoska2022}
\begin{align}
\label{eq:CondExpectRatioEstimateKDE}
    \widehat{\mathbb{E}[Y \mid g(X)]} := \frac{\sum_{j=1}^n k(g(X),g(x_j))y_j}{\sum_{j=1}^n k(g(X),g(x_j))},
\end{align}
where a natural choice for $k$ can be the Dirichlet kernel \citep{popordanoska2022}, defined as
\begin{equation}
    k_{Dir}(g(x_i),g(x_j)) = \frac{\Gamma(\sum_{k=1}^K \alpha_{jk})}{\prod_{k=1}^K \Gamma(\alpha_{jk})} \prod_{k=1}^K g(x_i)_{k}^{\alpha_{jk}-1}
\end{equation}
with $\alpha_{j} = \frac{g(x_j)}{h} + 1$ \citep{ouimet2020}. 
This results in a differentiable, asymptotically unbiased and consistent estimator of the conditional expectation. 

Another common choice for estimating the conditional expectation is by using a binning kernel (which returns 1 if $g(x_i)$ and $g(x_j)$ fall in the same bin, and 0 otherwise), resulting in an estimator given by $\widehat{\mathbb{E}[Y \mid g(X)]} = \frac{1}{|B_{g(x_h)}|} \sum_{j\in B_{g(x_h)}} y_j$,
where $B_{g(x_h)}$ denotes the bin into which $g(x_h)$ is assigned. Although this approach allows for faster computation, the estimator is not differentiable, thus preventing it to be directly used as part of calibration regularized training or differentiable recalibration methods. In addition, several papers \citep{vaicenavicius2019evaluating,widmann2019calibration,ashukha2021pitfalls} have raised concerns about asymptotic inconsistency of binned estimators, as well as sensitivity to the binning scheme, and limited scalability with the number of classes. 

\section{Statistical properties}

We show here the existence of consistent and asymptotically unbiased estimators of arbitrary proper calibration errors.
We first show in Section~\ref{sec:RefinementSameRateAsCE} that the optimal big-O convergence rates for estimators of $\operatorname{CE}_F$ and $\operatorname{REF}_F$ are the same and that a consistent estimator of one can be used to construct an estimator of the other.  We then prove in Section~\ref{sec:BregmanBiasRate} via a Taylor series approach that an estimator via $\operatorname{REF}_F$ gives asymptotic unbiasedness for \emph{all} Bregman divergences, giving a constructive proof that a single software implementation can provide a good baseline estimator for any proper calibration error using a function reference for $F$ to parameterize the Bregman divergence.

\subsection{Estimation via refinement}\label{sec:RefinementSameRateAsCE}

Equation~\eqref{eq:BregmanCalibrationError} provides a calibration error estimate directly from the definition of a Bregman divergence, but we show here that we can also use an estimate based on refinement using the decomposition (cf.\ Equation~\eqref{eq:decomposition_risk_ce_refinement})
\begin{align}  
\mathbb{E}
\left[D_{F}(Y,g(X))\right]
- \operatorname{CE}_{F}(g) = \mathbb{E} [ F(Y)  - 
    F( \mathbb{E}[Y \mid g(X)])] .
\end{align}
From this decomposition
and that $\mathbb{E}\left[D_{F}(Y,g(X))\right]$ can be estimated with an empirical average achieving an unbiased estimator with $\mathcal{O}(n^{-1/2})$ rate of convergence, we see that the difficulty of estimating refinement or CE is essentially the same, as the rate of bias and convergence for an estimator of one can be transferred to the other by subtracting from the empirical estimate of the risk.  Furthermore, we note that a simple empirical mean over $\{F(y_i)\}_{1\leq i\leq n}$ is the Minimum-Variance Unbiased Estimator (MVUE) of $\mathbb{E}[F(Y)]$ for a finite sample, and it is the estimate of $-\mathbb{E}[F(\mathbb{E}[Y\mid g(X)])]$ that is the primary challenge.

To summarize, assuming an empirical estimator of the refinement (cf.\ Equation~\eqref{eq:BregmanSharpnessEmpirical}) $\widehat{\operatorname{REF}}_F(g) \approx \mathbb{E}[-F(\mathbb{E}[Y\mid g(X)])]$, we may compute
\begin{align}
    \label{eq:BregmanCE_sharpness}\widehat{\operatorname{CE}}_F(g) := -\widehat{\operatorname{REF}}_F(g) + \frac{1}{n} \sum_{i=1}^n \left( D_F(y_i,g(x_i)) - F(y_i) \right)  ,
\end{align}
and the rates of convergence of $\widehat{\operatorname{CE}}_F(g)$ and its bias will be determined by the rates of $\widehat{\operatorname{REF}}_F(g)$.

\subsection{Asymptotic unbiasedness and rate of bias}\label{sec:BregmanBiasRate}

If we use the empirical estimator of $\mathbb{E}[Y\mid g(X)]$ in Equation~\eqref{eq:CondExpectRatioEstimateKDE}, the bias converges as $\mathcal{O}(n^{-1})$ while the estimator itself has a rate of $\mathcal{O}(n^{-1/2})$.
By the same argument as \citet[Footnote~2]{gruber2022better}, 
\begin{align}\label{eq:BregmanSharpnessEmpirical}
    \widehat{\operatorname{REF}}_F(g) := -\frac{1}{n} \sum_{h=1}^n F\left(\frac{\sum_{j\neq h} k(g(x_h),g(x_j))y_j}{\sum_{j\neq h} k(g(x_h),g(x_j))}\right)
\end{align}
is a consistent and asymptotically unbiased estimator of the refinement $\mathbb{E}[-F(\mathbb{E}[Y\mid g(X)])]$ for all $F$.  We note that continuity of $F$, a condition required for the argument of \citet[Footnote~2]{gruber2022better}, is guaranteed for all Bregman divergences as $F$ is differentiable by assumption.

We now show the asymptotic rate of bias of $\widehat{\operatorname{REF}}_F(g)$ using a Taylor series expansion.
First, define $\Delta = \widehat{\mathbb{E}[Y \mid g(x_h)]} - \mathbb{E}[Y \mid g(x_h)]$.
The rate of bias of $\widehat{\operatorname{REF}}_F(g)$ is determined by the rate of bias of each of the summands in Equation~\eqref{eq:BregmanSharpnessEmpirical}
\begin{align}
    &\mathbb{E}\left[F\left(\widehat{\mathbb{E}\left[Y \mid g(x_h)\right]}\right)\right] = \mathbb{E}\left[F\left(\mathbb{E}[Y \mid g(x_h)] + \Delta \right)\right] \notag \\
    &\approx  
    F\left(\mathbb{E}[Y \mid g(x_h)]\right) + \mathbb{E}\left[\operatorname{D}F\left(\mathbb{E}[Y \mid g(x_h)]\right)\Delta \right] \notag \\
    & \quad + \mathbb{E}\left[  \frac{1}{2} \Delta^T \operatorname{D}^2 F\left(\mathbb{E}[Y \mid g(x_h)]\right)\Delta\right] + \dots  
\end{align}
where $\operatorname{D}$ is the differential operator.
It is well known that the bias of the 1st order term (a ratio estimator) is $\mathcal{O}(n^{-1})$
and the remaining bias will be dominated by the 2nd order term \citep[Theorem~6.2.5]{wolter2007introduction}.  When $\operatorname{D}^2 F\left(\mathbb{E}[Y \mid g(x_h)]\right)\neq 0$, this yields
\begin{equation}
\begin{split}
    & \mathbb{E}\left[  
    \Delta^T \operatorname{D}^2 F\left(\mathbb{E}[Y \mid g(x_h)]\right)\Delta \right]  \\
    & \asymp \operatorname{Trace}\left[ \operatorname{Cov}\left( \widehat{\mathbb{E}[Y \mid g(x_h)]} \right) \right] 
    \\
    & \quad + \underbrace{\left\| \mathbb{E}\left[\widehat{\mathbb{E}[Y \mid g(x_h)]}\right] - \mathbb{E}[Y \mid g(x_h)]\right\|^2}_{=\left\|\operatorname{Bias}\left(\widehat{\mathbb{E}[Y \mid g(x_h)]}\right)\right\|^2 = \mathcal{O}(n^{-2})}, 
\end{split}
\end{equation}
where the notation $\asymp$ is taken here to mean that the left and right side have the same asymptotic rate of convergence in $n$, and the r.h.s.\ is due to a bias-variance decomposition of the l.h.s. Finally, we have
$\operatorname{Var}\left(   \widehat{\mathbb{E}[Y \mid g(x_h)]}_i 
    \right) = \mathcal{O}(n^{-1})$, which implies
\begin{align}
     \left| \operatorname{Bias}\left( \widehat{\operatorname{REF}}_F(g) \right) \right| =\mathcal{O}(n^{-1})
\end{align}
irrespective of $F$.  We therefore conclude that estimation of any proper calibration error via Equation~\eqref{eq:BregmanCE_sharpness} results in a consistent and asymptotically unbiased estimator with convergence $\mathcal{O}(n^{-1/2})$, and bias that converges as $\mathcal{O}(n^{-1})$.

\section{Relationship with information monotonicity in neural networks}

A key part of this work is to relate uncertainty calibration to information theoretic principles.
First, we summarize relevant concepts and provide the necessary foundation to derive our contributions.

\subsection{Background on information monotonicity}
In machine learning, information monotonicity is most commonly known through the information bottleneck theory \citep{slonim1999agglomerative, bialek2001predictability, gilad2003information, chechik2003information, shamir2010learning, shwartz2017opening, michael2018on}.
Information monotonicity states that each layer in a neural network is indirectly optimized by the mutual information of the output, resulting in a so-called information flow, or information plane dynamics, throughout the network \citep{michael2018on, goldfeld2019estimating}.

Further, \citet{csiszar1972class} derived the class of f-divergences according to several principles, including information monotonicity.
These divergences between distributions are widely applicable, for example throughout statistics \citep{liese2006divergences} and in generative modelling \citep{creswell2018generative}.
Similar to \citet{garcia2012divergences} and \citet{duchi2018multiclass}, we use the following definition for multiple distributions.
Given a convex function $f \colon \left[0, \infty \right)^k \to \left( - \infty, \infty \right]$ with $f \left( 1, \dots, 1 \right) = 0$, the \textbf{f-divergence} between distributions $P_1, \dots, P_k$ and $Q$ is defined by
\begin{equation}
    I_f \left( P_1, \dots, P_k \parallel Q \right) = \int f \left( \frac{\mathrm{d}P_1}{\mathrm{d}Q}, \dots, \frac{\mathrm{d}P_k}{\mathrm{d}Q} \right) \mathrm{d}Q.
\end{equation}
Following the property of $f$, we have $I_f \left( P_1, \dots, P_k \parallel Q \right) \geq 0$ with equality if $P_1 = \dots = P_k = Q$.
Let $M$ be a Markov kernel transforming a distribution $P$ into a distribution $MP$, then \citet{garcia2012divergences} show that the information monotonicity is given by
\begin{equation}
    I_f \left( M P_1, \dots, M P_k \parallel M Q \right) \leq I_f \left( P_1, \dots, P_k \parallel Q \right).
\end{equation}

In the following, we make the novel connection between f-Divergences and refinement via model sharpness.
The sharpness of a model is defined via \citep{degroot1981assessing}
\begin{equation}
    \operatorname{SHARP}_F \left( g \right) = \mathbb{E} \left[ D_F \left( \mathbb{E} \left[ Y \mid g \left( X \right) \right], \mathbb{E} \left[ Y \right] \right) \right].
\end{equation}
If $F$ is the Shannon entropy, then the sharpness is equivalent to the mutual information between output variable and target variable.
Model sharpness is related to the refinement term \citep{degroot1981assessing, morris1983comparison, kull2015novel, kuleshov2022calibrated} by
\begin{equation}\label{eq:SharpnessRefinement}
    - \operatorname{SHARP}_F \left( g \right) = F \left( \mathbb{E} \left[ Y \right] \right) + \underbrace{\mathbb{E} \left[ - F \left( \mathbb{E} \left[ Y \mid g \left( X \right) \right] \right) \right]}_{\text{ (Refinement)}}.
\end{equation}
Consequently, we can use our proposed refinement estimator for calibration and sharpness.

\subsection{A novel generalization of neural network information monotonicity}
\label{sec:novel_info_mono}
We now present our contribution regarding the existence of information monotonicity in neural networks.
As a preliminary step, we first show that model sharpness has the form of an f-divergence.
We defer proofs to Appendix \ref{appendix:missing_proofs_inf_mono}.

\begin{proposition}[Sharpness as f-divergence]\label{prop:sharp_f_div}
Let $F \colon \Delta^k \to \mathbb{R}$ be a convex function and $g \colon \mathcal{X} \to \Delta^k$ a classifier with prediction distributions $P_y \coloneqq \mathbb{P} \left( g \left( X \right) \mid Y = y \right)$, and $P \coloneqq \mathbb{P} \left( g \left( X \right) \right)$. 
Then, the model sharpness can be represented as an f-divergence via
\begin{equation}\operatorname{SHARP}_F \left( g \right) = I_{F^Y} \left( P_1, \dots, P_k \parallel P \right),
\end{equation}

where $F^Y \left( x \right) \coloneqq F \left( \mathbb{E} \left[ Y_1 \right] x_1, \dots, \mathbb{E} \left[ Y_k \right] x_k \right) - F \left( \mathbb{E} \left[ Y_1 \right], \dots, \mathbb{E} \left[ Y_k \right] \right)$.
\end{proposition}

Thus, we can interpret the classifier sharpness as the $f$-Divergence between the class-conditional prediction distributions and the marginal prediction distribution.
The marginal class distribution determines the weight of each ratio.
For a given classification task, it is constant across all models.
We can now provide the key result of this section.

\begin{theorem}[Information monotonicity in neural networks]\label{th:imnn}

For a neural network $g \left( X \right) = h_l \left( \dots \left( h_1 \left( X \right) \right) \right)$ with layers $h_i$, $i \in \left\{ 1, \dots, l \right\}$, conditional distributions $P^i_y \coloneqq \mathbb{P} \left( h_i \left( \dots \left( h_1 \left( X \right) \right) \right) \mid Y = y \right)$, and marginal distributions $P^{i} \coloneqq \mathbb{E} \left[ P^{i}_Y \right]$, we have
\begin{equation}
\begin{split}    
    \operatorname{SHARP}_F \left( g \right) & = I_{F^Y} \left( P^{{l}}_1, \dots, P^{{l}}_k \parallel P^{{l}} \right) \\
    & \leq I_{F^Y} \left( P^{{l-1}}_1, \dots, P^{{l-1}}_k \parallel P^{{l-1}} \right) \\
    & \leq \dots \leq I_{F^Y} \left( P^{1}_1, \dots, P^{1}_k \parallel P^{1} \right). \\
\end{split}
\end{equation}
\end{theorem}
This theorem offers a generalization of the information flow in neural networks.
Since sharpness is optimized towards the maximum during model training, the information (as quantified by an f-divergence) is implicitly maximized in each layer no matter the proper loss.
The signal, which is forward propagated in each layer, follows the known information flow of the information bottleneck theory.
A rich literature exists on information bottleneck experiments \citep{shwartz2017opening, michael2018on, goldfeld2019estimating, Wu2020Phase, wu2020learnability, wang2022pacbayes}. 
In Section \ref{sec:experiments} we will extend on that by monitoring the model sharpness via the refinement term throughout model training and by assessing information monotonicity beyond mutual information.

In Section \ref{sec:background}, we have seen that calibration and sharpness are two different yet related concepts derived from risk minimization.
Theorem \ref{th:imnn} presents a novel link between information bottleneck theory and uncertainty calibration -- two key research areas in deep learning, which consist of rich literature but little exchange.
For example, according to our result, optimizing via the information bottleneck theory does not offer calibrated predictions and requires post-hoc uncertainty calibration for trustworthy probability forecasts.
This underlines the general importance of calibration estimation.

\section{Experiments}
\label{sec:experiments}

\begin{figure*}[ht]
    \centering
    \subfloat[Dirichlet vs. Binning]{
    \resizebox{0.32\textwidth}{!}{\input{figs/bin_vs_kde_kl}}
    \label{subfig:bin_vs_kde_kl}
    }
    \subfloat[Direct vs. estimation via risk]{
    \resizebox{0.32\textwidth}{!}
    {\input{figs/two_impl_kl}}
    \label{subfig:two_impl_KL}
    }
    \subfloat[Bias convergence]{
    \resizebox{0.32\textwidth}{!}{\input{figs/bias_convergence_synthetic_kl}}
    \label{subfig:bias_convergence_synthetic_kl}
    }
    \caption{\subref{subfig:bin_vs_kde_kl} A comparison of the binning-based estimator and the KDE-based estimator with Dirichelt kernel. \subref{subfig:two_impl_KL} A comparison of the CE estimator using Equation~\eqref{eq:BregmanCE_sharpness} and direct estimation via Equation~\eqref{eq:BregmanCalibrationError}. \subref{subfig:bias_convergence_synthetic_kl} Convergence of the bias as a function of the number of points used for the estimation. }
    \label{fig:synthetic_data}
\end{figure*}

The outline of the experiments is as follows. We 
\begin{enumerate*}[label=(\roman*)]
\item compare the choices of kernel for the conditional expectation estimator, discussed in Section \ref{sec:estimating_proper_ce};
\item compare the direct estimator from Equation~\eqref{eq:BregmanCalibrationError} with the estimator derived via risk in Equation~\eqref{eq:BregmanCE_sharpness};
\item analyse the empirical properties of the proposed estimator, derived in Section \ref{sec:RefinementSameRateAsCE};
\item show new insights regarding the choice of post-hoc calibration method, depending on the chosen calibration error;
\item demonstrate the information monotonicity in neural networks, discussed in Section \ref{sec:novel_info_mono}, for the $L_2$ case via our proposed estimator.
\end{enumerate*}

In all experiments we evaluate class-wise CE, which can be derived from One-vs-Rest risk minimization (c.f. Appendix \ref{appendix:classwise_calibration_one_v_rest}). The CE estimator obtained by setting $F(x) = \|x\|_2^2$ in Equation~\ref{eq:BregmanCE_sharpness} will be denoted as $\widehat{\operatorname{CE}_{2}^{2}}$, while the one derived from $F(p) = \langle p, \log(p)\rangle$ will be referred to as $\widehat{\operatorname{CE}}_{\mathrm{KL}}$.
The bandwidth of the Dirichlet kernel is determined through a combination of a leave-one-out maximum likelihood estimation and visual inspection of the resulting density. Typical values range from $0.01$ to $0.0001$.

\subsection{Empirical properties}

To analyze the empirical properties of the proposed estimator, we create synthetic data with miscalibrated scores, for which the ground truth CE is known, following \citet{popordanoska2022}. First, we sample uniform points from the simplex and we apply temperature scaling with $t_1 = 0.9$ to ensure that the scores are closer to the boundaries of the simplex. Then, we generate ground truth labels based on the sampled probabilities, resulting in a perfectly calibrated classifier. Finally, to intentionally introduce miscalibration, we apply an additional temperature scaling with $t_2 = 0.6$. 

In Figure~\ref{subfig:bin_vs_kde_kl} we compare the performance of two proposed choices of kernel, $k_{Dir}$ and $k_{bin}$, for increasing number of samples used for the estimation. We observe that the Dirichlet-based estimator not only has better properties, like differentiability, consistency and asymptotic unbiasedness, but also has better empirical performance.
Figure~\ref{subfig:two_impl_KL} compares the direct implementation of the estimator, as given in Equation~\eqref{eq:BregmanCalibrationError}, with the estimator derived via the risk, given in Equation~\eqref{eq:BregmanCE_sharpness}. We derived theoretically the statistical properties of the estimator in Equation~\eqref{eq:BregmanCE_sharpness} (blue curve), and empirically the direct implementation (orange) performs even better.  
Finally, Figure~\ref{subfig:bias_convergence_synthetic_kl} shows the convergence of the bias of $\widehat{\operatorname{CE}}_{\mathrm{KL}}$ as a function of the number of points used for the estimation. We observe that regardless of the number of classes, the estimator consistently provides reliable estimates of class-wise calibration error.

\subsection{Post-hoc calibration}

\begin{table*}[ht]
    \centering
    \caption{Performance evaluation ($\widehat{\operatorname{CE}}_{\mathrm{KL}} \times 100$ and $\widehat{\operatorname{CE}_{2}^{2}} \times 100$, lower is better) of various network architectures on CIFAR-10/100 with no calibration, and after recalibrating with IR and TS. The number in the bracket represents the change of CE (in $\%$) relative to the uncalibrated score. The results are averaged over 5 seeds.}
    \label{tab:calibration_methods_ce}
\resizebox{1\textwidth}{!}{
    \begin{tabular}{cccccccc}
        \toprule
        & & \multicolumn{2}{c}{\textbf{No calibration}} & \multicolumn{2}{c}{\textbf{Isotonic regression}} & \multicolumn{2}{c}{\textbf{Temperature scaling}} \\
        Dataset & Model & $\widehat{\operatorname{CE}}_{\mathrm{KL}}$ & $\widehat{\operatorname{CE}_{2}^{2}}$ & $\widehat{\operatorname{CE}}_{\mathrm{KL}}$ & $\widehat{\operatorname{CE}_{2}^{2}}$ & $\widehat{\operatorname{CE}}_{\mathrm{KL}}$ & $\widehat{\operatorname{CE}_{2}^{2}}$ \\
        \midrule
        \multirow{6}{*}{CIFAR-10} 
        & PreResNet20 & \cellcolor{graygray}{$2.74_{\pm 0.06}$} & $1.12_{\pm 0.01}$ & \cellcolor{graygray}{$1.57_{\pm 0.05} (\downarrow 43 \%)$} & $0.58_{\pm 0.01} (\downarrow 48 \%)$ & \cellcolor{graygray}{$1.26_{\pm 0.02} (\downarrow 54 \%)$} & $0.66_{\pm 0.01} (\downarrow 41 \%)$ \\ 
        & PreResNet56 & \cellcolor{graygray}{$1.94_{\pm 0.04}$} & $0.84_{\pm 0.03}$ & \cellcolor{graygray}{$1.30_{\pm 0.03} (\downarrow 33 \%)$} & $0.50_{\pm 0.03} (\downarrow 41 \%)$ & \cellcolor{graygray}{$1.03_{\pm 0.02} (\downarrow 47 \%)$}& $0.54_{\pm 0.02} (\downarrow 36 \%)$ \\ 
        & PreResNet110 & \cellcolor{graygray}{$1.76_{\pm 0.03}$} & $0.77_{\pm 0.01}$ & \cellcolor{graygray}{$1.26_{\pm 0.03} (\downarrow 29 \%)$} & $0.49_{\pm 0.01} (\downarrow 37 \%)$ & \cellcolor{graygray}{$0.98_{\pm 0.01} (\downarrow 44 \%)$} & $0.51_{\pm 0.01} (\downarrow 33 \%)$ \\ 
        & PreResNet164 & \cellcolor{graygray}{$1.58_{\pm 0.02}$} & $0.70_{\pm 0.01}$ & \cellcolor{graygray}{$1.20_{\pm 0.02} (\downarrow 24 \%)$} & $0.43_{\pm 0.02} (\downarrow 38 \%)$ & \cellcolor{graygray}{$0.91_{\pm 0.02} (\downarrow 42 \%)$}& $0.47_{\pm 0.01} (\downarrow 32 \%)$ \\ 
        & VGG16BN & \cellcolor{graygray}{$2.85_{\pm 0.05}$} & $1.00_{\pm 0.02}$ & \cellcolor{graygray}{$1.36_{\pm 0.03} (\downarrow 52 \%)$} & $0.57_{\pm 0.01} (\downarrow 42 \%)$ & \cellcolor{graygray}{$1.38_{\pm 0.03} (\downarrow 52 \%)$} & $0.80_{\pm 0.02} (\downarrow 20 \%)$ \\ 
        & WideResNet28x10 & \cellcolor{graygray}{$1.21_{\pm 0.02}$} & $0.61_{\pm 0.01}$ & \cellcolor{graygray}{$1.09_{\pm 0.03} (\downarrow 10 \%)$} & $0.41_{\pm 0.01} (\downarrow 34 \%)$ & \cellcolor{graygray}{$0.93_{\pm 0.01} (\downarrow 23 \%)$} & $0.49_{\pm 0.01} (\downarrow 19 \%)$ \\ 
        \midrule
        \multirow{6}{*}{CIFAR-100} 
        & PreResNet20 & \cellcolor{graygray}{$0.76_{\pm 0.01}$} & $0.38_{\pm 0.00}$ & \cellcolor{graygray}{$0.96_{\pm 0.02} (\uparrow 26 \%)$} & $0.35_{\pm 0.00} (\downarrow 9 \%)$ & \cellcolor{graygray}{$0.70_{\pm 0.00} (\downarrow 8 \%)$} & $0.35_{\pm 0.00} (\downarrow 8 \%)$ \\ 
        & PreResNet56 & \cellcolor{graygray}{$0.78_{\pm 0.02}$} & $0.35_{\pm 0.01}$ & \cellcolor{graygray}{$0.88_{\pm 0.01} (\uparrow 13 \%)$} & $0.28_{\pm 0.00} (\downarrow 21 \%)$ & \cellcolor{graygray}{$0.61_{\pm 0.01} (\downarrow 22 \%)$} & $0.30_{\pm 0.00} (\downarrow 14 \%)$ \\ 
        & PreResNet110 & \cellcolor{graygray}{$0.76_{\pm 0.01}$} & $0.34_{\pm 0.00}$ & \cellcolor{graygray}{$0.85_{\pm 0.01} (\uparrow 12 \%)$} & $0.27_{\pm 0.00} (\downarrow 20 \%)$ & \cellcolor{graygray}{$0.60_{\pm 0.00} (\downarrow 21 \%)$} & $0.30_{\pm 0.00} (\downarrow 12 \%)$ \\ 
        & PreResNet164 & \cellcolor{graygray}{$0.74_{\pm 0.00}$} & $0.33_{\pm 0.00}$ & \cellcolor{graygray}{$0.86_{\pm 0.01} (\uparrow 16 \%)$} & $0.26_{\pm 0.00} (\downarrow 21 \%)$ & \cellcolor{graygray}{$0.59_{\pm 0.01} (\downarrow 20 \%)$} & $0.29_{\pm 0.00} (\downarrow 11 \%)$ \\ 
        & VGG16BN & \cellcolor{graygray}{$1.23_{\pm 0.01}$} & $0.43_{\pm 0.00}$ & \cellcolor{graygray}{$0.95_{\pm 0.01} (\downarrow 23 \%)$} & $0.34_{\pm 0.00} (\downarrow 21 \%)$ & \cellcolor{graygray}{$0.75_{\pm 0.00} (\downarrow 39 \%)$} & $0.38_{\pm 0.00} (\downarrow 11 \%)$ \\ 
        & WideResNet28x10 & \cellcolor{graygray}{$0.62_{\pm 0.01}$} & $0.30_{\pm 0.00}$ & \cellcolor{graygray}{$0.72_{\pm 0.02} (\uparrow 15 \%)$} & $0.22_{\pm 0.00} (\downarrow 27 \%)$ & \cellcolor{graygray}{$0.60_{\pm 0.01} (\downarrow 3 \%)$} & $0.30_{\pm 0.00} (\downarrow 1 \%)$ \\ 
        \bottomrule
    \end{tabular}
    }
\end{table*}  

\begin{table}[ht]
    \centering
    \caption{Accuracy on CIFAR-10.}
    \resizebox{1\columnwidth}{!}{
    \begin{tabular}{ccc}
        \toprule
        Model & No calibration & Isotonic regression \\
        \midrule
         PreResNet20 & $91.95_{\pm 0.05}$ & $91.94_{\pm 0.07}$ \\
         PreResNet56 & $94.38_{\pm 0.13}$ & $94.34_{\pm 0.13}$ \\
         PreResNet110 & $94.86_{\pm 0.04}$ & $94.83_{\pm 0.05}$ \\
         PreResNet164 & $95.24_{\pm 0.05}$ & $95.14_{\pm 0.06}$\\
         VGG16BN & $93.26_{\pm 0.04}$ & $93.23_{\pm 0.04}$ \\
         WideResNet28x10 & $95.54_{\pm 0.05}$ & $95.53_{\pm 0.04}$ \\
        \bottomrule
    \end{tabular}
    }
    \label{tab:accuracy_cifar_10}
\end{table}

Here we demonstrate the application of our proposed estimator for evaluating CE on CIFAR10/100~\citep{krizhevsky2009learning}, after performing post-hoc calibration. Following standard practice, we trained various PreResNet ~\citep{he2016identity}, VGG16~\citep{simonyan2014very} and WideResNet~\citep{zagoruyko2016wide} architectures.
Details about the training can be found in Appendix \ref{appendix:experiments}.

In Table~\ref{tab:calibration_methods_ce} we present a comparison of $\widehat{\operatorname{CE}}_{\mathrm{KL}}$ and $\widehat{\operatorname{CE}_{2}^{2}}$ for models before and after calibration with temperature scaling (TS) and isotonic regression (IR) \citep{guo2017calibration}. We focus on these methods because of their distinct optimization objectives during calibration: TS minimizes the NLL loss, while IR aims to optimize a weighted Brier score. It is notable that we obtain a much better $\widehat{\operatorname{CE}}_{\mathrm{KL}}$ score with TS across all architectures on both datasets. For instance, we report $42\%$ improvement from the uncalibrated score using TS, compared with $24\%$ decrease in CE using IR for PreResNet164 on CIFAR-10. The effect is opposite for $\widehat{\operatorname{CE}_{2}^{2}}$: IR is a better suited calibration technique if one aims to minimize this metric. For example, calibration with IR on VGG16BN results in 42$\%$ decrease in CE, compared to only 20$\%$ decrease with TS. On the other hand, if the goal is to optimize $\widehat{\operatorname{CE}}_{\mathrm{KL}}$, the results on CIFAR-100 indicate that IR may even harm this metric. Table~\ref{tab:accuracy_cifar_10} summarizes the accuracy on CIFAR-10 before and after calibration with IR. We notice that while TS is known to be accuracy-perserving, IR also retains accuracy in practice for this setting. 
In summary, these findings suggest that the choice of calibration method should be influenced by the specific calibration error of interest, i.e., IR is more suitable for minimizing $\widehat{\operatorname{CE}_{2}^{2}}$, whereas TS should be preferred for $\widehat{\operatorname{CE}}_{\mathrm{KL}}$.

The full table evaluating accuracy, NLL and Brier score is in Appendix \ref{appendix:experiments}.
Additional experiments involving convergence of bias, monitoring CE and sharpness during training, performing model selection, and measuring class-wise CE are also discussed in that Appendix.

\subsection{Information monotonicity}

We illustrate the information monotonicity in neural networks via our refinement estimator.
We can apply our refinement estimator also to the sharpness of random vectors not located in the simplex.
For this, note that $\mathbb{E} \left[ f \left( \mathbb{E} \left[ Y \mid X \right] \right) \right] = \mathbb{E} \left[ f \left( \mathbb{E} \left[ Y \mid g \left( X \right) \right] \right) \right]$ for any function $f$ and injective function $g$ (c.f. Appendix D.8 in \citep{gruber2022better}).
In our case, $f$ is the convex function of a refinement term and $g$ is chosen to be an invertible version of the softmax function and $X$ are the activations of an intermediate layer. \\

\begin{figure}[h]
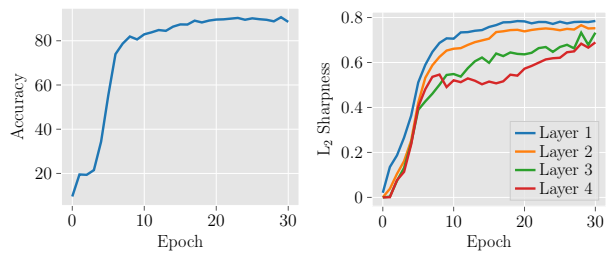

    \centering
    \resizebox{0.48\columnwidth}{!}{\input{figs/info_mono_plot_acc}}
    \resizebox{0.48\columnwidth}{!}{\input{figs/info_mono_plot_l2}}
    \caption{Accuracy and information monotonicity throughout training via our sharpness estimator (here: L$_2$). Each consecutive layer's sharpness is lower bounded by the next layer. The sharpness of the whole model is optimized implicitly via a proper score.}
    \label{fig:info_mono}
\end{figure}

We train a fully connected neural network on MNIST via stochastic gradient descent.
The model has four hidden layers with nine nodes each.
In Figure \ref{fig:info_mono}, we show that the model accuracy and $L_2$ sharpness of each intermediate layer throughout training.
Note that sharpness can also be interpreted as mutual information.
At the beginning of training, the information in each layer increases sharply similar to the accuracy.
But, the initial increase of information holds no longer for layers one and two, while layers three and four experience a slow-down in information gain.
This information gap is then reduced for longer periods of training.
We conclude that the information in each layer is not optimized uniformly throughout training, which can be monitored via our refinement estimator.

\section{Discussion and conclusions}

In this paper, we proposed a consistent and differentiable estimator for all proper calibration errors. The estimator is asymptotically unbiased, has a convergence rate of $\mathcal{O}(n^{-1/2})$, and the bias decreases at a rate of $\mathcal{O}(n^{-1})$.
Specifically, we introduce the Kullback--Leibler CE as a theoretically justified choice in standard neural network training procedures that utilize log loss. 
Furthermore, we showed that model sharpness, a generalization of mutual information, is equal to a multi-distribution f-divergence.
Via this relation, we proved that information monotonicity in neural networks is a general concept beyond log minimization and can be monitored via sharpness during model training.

Our work has several limitations. It is an open research question to find bias corrections for proper CE estimators. Although \citet{popordanoska2022} provided a debiasing strategy for $\widehat{\operatorname{CE}_{2}^{2}}$, it does not transfer directly to our more general case.
Further, an intrinsic problem of density estimators for CE is the $\mathcal{O}(n^2)$ complexity. In our experiments, we demonstrate that evaluation can be effectively computed on subsets of reasonable size. However, assessing larger sets becomes computationally expensive. Improved debiasing could make block estimators feasible \citep{NIPS2013_a49e9411}, leading also to improved computational properties.

In summary, we show that there exist asymptotically unbiased estimators for an entire landscape of proper CEs.
The experimental results demonstrate the empirical behavior of the proposed approach and showcase its properties in assessing CE and sharpness. 
This makes it a valuable component for designing calibration methods which aim to minimize a specific CE.

\section*{Acknowledgements}
This research received funding from the Flemish Government (AI Research Program) and the Research Foundation - Flanders (FWO) through project number G0G2921N. 
The publication was also supported by funding from the Academy of Finland (Profi6 336449 funding program), the University of Oulu strategic funds, the Wellbeing Services County of North Ostrobothnia (VTR project K62716), Terttu Foundation, and the Finnish Foundation for Cardiovascular Research. The authors wish to acknowledge CSC -- IT Center for Science, Finland, for generous computational resources.\\
Co-funded by the European Union (ERC, TAIPO, 101088594 to FB). Views and opinions expressed are however those of the authors only and do not necessarily reflect those of the European Union or the European Research Council. Neither the European Union nor the granting authority can be held responsible for them.

\clearpage

{
\small

\bibliography{refs}
\bibliographystyle{icml2022}
}

\appendix
\onecolumn
\section{Uncertainty calibration in classification}
\label{appendix:uncertainty_calibration_in_classification}

In this section, we give a more detailed overview of calibration errors compared to the main paper.
To introduce calibration formally, we consider a classifier $g \colon \mathcal{X} \to \simplex{k}$, where $\simplex{k} \coloneqq \left\{ \left(p_1, \dots, p_k \right)^\intercal \in \left[0, 1 \right]^k \mid \sum_{i=1}^k p_i = 1 \right\}$ is a probability simplex with $k$ vertices, and $ \mathcal{X}$ is a feature space with feature variable $X$.
We denote class labels as one-hot encoded variables of $\mathcal{Y}=\{1, \dots, k\}$, i.e. the target variable $Y$ has observations $y \in \{0, 1\}^k \subset \simplex{k}$ with $\left\lVert y \right\rVert_1=1$. In the literature, there exist multiple notions of calibration of different strength~\citep{vaicenavicius2019evaluating, kull2019beyond}. 
Calibration errors assess the degree of violation of a respective notion.

The strongest notion of calibration is the canonical calibration, which compares the probability vector prediction $g \left( X \right)$ with the target distribution $\mathbb{E} \left[ Y \mid g \left( X \right) \right]$ given this prediction.
A class of canonical calibration errors, which was studied recently in several works~\citep{naeini2015,kumar2019verified,wenger2020,popordanoska2022, gruber2022better}, is referred to as $L_p$ calibration error and defined as
\begin{align}\label{eq:lp_canonical_cal}
    \operatorname{CE}_p(f) = \left(\expectation \left[\left\lVert\expectation \left[Y \mid g(X) \right] - g(X) \right\rVert_p^p\right] \right)^{\frac{1}{p}}.
\end{align}
The special case $\widehat{\operatorname{CE}_{2}^{2}}$, as given in Equation \eqref{eq:L2calibrationError}, can be derived from the Brier score \citep{ANewVectorPartitionoftheProbabilityScore}.

Canonical calibration errors are notoriously difficult to estimate and represent a calibration strictness which may not be necessary in practice \citep{vaicenavicius2019evaluating}.
One common and less constraining notion is class-wise calibration, which compares the individual prediction $g_i \left( X \right)$ of class $i \in \mathcal{Y}$ with the target class distribution $\mathbb{P} \left( Y = i \mid g_i \left( X \right) \right)$ given the individual class prediction.
The class-wise calibration error with respect to a given $L_p$ space is defined as
\begin{align}
    \operatorname{CWCE}_p \left( g \right) = \left( \frac{1}{k} \sum_{i=1}^k \expectation \left[ \left( \mathbb{P} \left(Y = i \mid g_i \left( X \right) \right) - g_i \left( X \right) \right)^p \right] \right)^\frac{1}{p}.
\end{align}
The definition is formalized based on \citet{kumar2019verified} and \citet{gruber2022better}, while the class-wise concept was introduced independently by \citet{kull2019beyond} and \citet{nixon2019measuring}.

While class-wise calibration is easier to evaluate than canonical calibration, it still scales linearly in complexity with the number of classes, which can be problematic for tasks with a very high number of classes.
In contrast, the most common approach in the machine learning literature is top-label confidence calibration \citep{naeini2015, guo2017calibration, Joo2020BeingBA, Kristiadi2020BeingBE, rahimi2020intra, Tomani_2021_CVPR, minderer2021revisiting, Tian2021AGP, Islam2021ABS, Menon2021ASP, Moraleslvarez2021ActivationlevelUI, Gupta2021BERTF, Wang2021BeCT, Fan2022AcceleratingBN}, which is not affected by this issue.
In this notion, we compare if the predicted top-label confidence $\max_{i \in \mathcal{Y}} g_i \left( X \right)$ matches the conditional accuracy $\mathbb{P} \left( Y = \arg\max_{i \in \mathcal{Y}} g_i \left( X \right) \mid \max_{i \in \mathcal{Y}} g_i \left( X \right) \right)$ given the predicted top-label confidence.
It is known in the literature that top-label confidence calibration represents the weakest notion of calibration~\citep{vaicenavicius2019evaluating, widmann2019calibration, gruber2022better}.
The top-label confidence calibration error based on an $L_p$ space is defined as~\citep{kumar2019verified, gruber2022better}

\begin{align}
    \operatorname{TCE}_p \left( g \right) = \left( \expectation \left[ \left( \mathbb{P} \left( Y = \arg\max_{j\in \mathcal{Y}} g_j \left( X \right) \mid \max_{j \in \mathcal{Y}} g_j \left( X \right) \right) - \max_{j \in \mathcal{Y}} g_j \left( X \right) \right)^p \right] \right)^\frac{1}{p}.
\end{align}
For $p=1$, its binning based estimator is commonly referred to as expected calibration error \citep{naeini2015, guo2017calibration}.

Besides the $L_p$ based calibration errors presented above, there also exist other calibration errors, like maximum mean calibration error \citep{pmlr-v80-kumar18a}, Kolmogorov-Smirnov calibration error \citep{gupta2020calibration}, and kernel calibration error \citep{widmann2019calibration}.
We exclude these errors from our analysis and experiments as they are not related to risk minimization.

\section{Proofs for information monotonicity}
\label{appendix:missing_proofs_inf_mono}

In this section, we provide detailed proofs about the f-divergence representation of the model sharpness and the information monotonicity in neural networks.
Since we will require the target variable $Y$ in categorical encoding and one-hot encoding, we will write $Y$ for the former and $\Vec{Y}$ for the latter.
This notation is exclusive to this section.

\subsection{Proof of Proposition \ref{prop:sharp_f_div}}
Let $F \colon \Delta^k \to \mathbb{R}$ be a convex function and $g \colon \mathcal{X} \to \Delta^k$ a classifier with prediction distributions $P_y \coloneqq \mathbb{P} \left( g \left( X \right) \mid Y = y \right)$, and $P \coloneqq \mathbb{P} \left( g \left( X \right) \right)$. 
Then, the model sharpness can be represented as an f-divergence via
\begin{equation}
\begin{split}
    & \operatorname{SHARP}_F \left( g \right) \\
    & = \mathbb{E} \left[ D_F \left( \mathbb{E} \left[ \Vec{Y} \mid g \left( X \right) \right], \mathbb{E} \left[ \Vec{Y} \right] \right) \right] \\
    \overset{\text{def}}&{=} \mathbb{E} \left[ F \left( \mathbb{E} \left[ \Vec{Y} \mid g \left( X \right) \right] \right) - F \left( \mathbb{E} \left[ \Vec{Y} \right] \right) - \left\langle \nabla F \left( \mathbb{E} \left[ \Vec{Y} \right] \right), \mathbb{E} \left[ \Vec{Y} \right] - \mathbb{E} \left[ \Vec{Y} \mid g \left( X \right) \right] \right\rangle \right] \\
    & = \mathbb{E} \left[ F \left( \mathbb{E} \left[ \Vec{Y} \mid g \left( X \right) \right] \right) \right] - F \left( \mathbb{E} \left[ \Vec{Y} \right] \right) \\
    & = \int_{\mathcal{X}} F \left( \mathbb{E} \left[ \Vec{Y} \mid g \left( X \right) \right] \right) - F \left( \mathbb{E} \left[ \Vec{Y} \right] \right) \mathrm{d} \mathbb{P} \left( g \left( X \right) \right) \\
    & = \int_{\mathcal{X}} F \left(\mathbb{E} \left[ \Vec{Y}_1 \right] \frac{\mathrm{d}\mathbb{P} \left( g \left( X \right) \mid Y = 1 \right)}{\mathrm{d} \mathbb{P} \left( g \left( X \right) \right)}, \dots, \mathbb{E} \left[ \Vec{Y}_k \right] \frac{\mathrm{d}\mathbb{P} \left( g \left( X \right) \mid Y = k \right)}{\mathrm{d} \mathbb{P} \left( g \left( X \right) \right)} \right) \\
    & \quad \quad - F \left( \mathbb{E} \left[ \Vec{Y} \right] \right) \mathrm{d} \mathbb{P} \left( g \left( X \right) \right) \\
    & = \int_{\mathcal{X}} F \left(\mathbb{E} \left[ \Vec{Y}_1 \right] \frac{\mathrm{d} P_1}{\mathrm{d} P}, \dots, \mathbb{E} \left[ \Vec{Y}_k \right] \frac{\mathrm{d} P_k}{\mathrm{d} P} \right) - F \left( \mathbb{E} \left[ \Vec{Y} \right] \right) \mathrm{d} P \\
    & = I_{F^Y} \left( P_1, \dots, P_k \parallel P \right) \\    
\end{split}
\end{equation}

where $F^Y \left( x \right) \coloneqq F \left( \mathbb{E} \left[ \Vec{Y}_1 \right] x_1, \dots, \mathbb{E} \left[ \Vec{Y}_k \right] x_k \right) - F \left( \mathbb{E} \left[ \Vec{Y}_1 \right], \dots, \mathbb{E} \left[ \Vec{Y}_k \right] \right)$ and by using Bayes' theorem.
The function $I_{F^Y}$ is a multi-distribution f-divergence since $F^Y$ is convex (follows from $F$ being convex) and $F^Y \left( 1, \dots, 1 \right) = F \left( \mathbb{E} \left[ \Vec{Y}_1 \right], \dots, \mathbb{E} \left[ \Vec{Y}_k \right] \right) - F \left( \mathbb{E} \left[ \Vec{Y}_1 \right], \dots, \mathbb{E} \left[ \Vec{Y}_k \right] \right) = 0$.

\subsection{Proof of Theorem \ref{th:imnn}}

For a neural network $g \left( X \right) = h_l \left( \dots \left( h_1 \left( X \right) \right) \right)$ with layers $h_i$, $i \in \left\{ 1, \dots, l \right\}$, conditional distributions $P^i_y \coloneqq \mathbb{P} \left( h_i \left( \dots \left( h_1 \left( X \right) \right) \right) \mid Y = y \right)$, and marginal distributions $P^{i} \coloneqq \mathbb{E} \left[ P^{i}_Y \right]$, we have
\begin{equation}
\begin{split}    
    \operatorname{SHARP}_F \left( g \right) & = I_{F^Y} \left( P^{{l}}_1, \dots, P^{{l}}_k \parallel P^{{l}} \right) \leq I_{F^Y} \left( P^{{l-1}}_1, \dots, P^{{l-1}}_k \parallel P^{{l-1}} \right) \\
    & \leq \dots \leq I_{F^Y} \left( P^{1}_1, \dots, P^{1}_k \parallel P^{1} \right) \leq \operatorname{SI}_F \left( X; Y \right). 
\end{split}
\end{equation}

\begin{proof}
    Since a neural network in our context is simply a chain of function, it is sufficient to prove that the inequality holds for a single arbitrary function $f$ transforming a sample space $\Omega$ with $\sigma$-field $\mathcal{F}$.
    For this, assume we are in the context of a probability space $\left( \Omega, \mathcal{F}, \mathbb{P} \right)$ and measurable space $\left( \Omega_f, \mathcal{F}_f \right)$ such that $f \colon \Omega \to \Omega_f$ is a measurable function.
    Define the function $M_f \colon \Omega \times \mathcal{F}_f \to \mathbb{R}$ via $M_f \left( \omega, A \right) = \mathbbm{1}_{\left\{ \omega \in \Omega \mid f \left( \omega \right) \in A \right\}} \left( \omega \right)$, 
    where $\mathbbm{1}$ is the indicator function.
    Similar as \citet{garcia2012divergences}, we write $M_f\mathbb{P} \left( A \right) = \int_\Omega M_f \left( \omega, A \right) \mathrm{d} \mathbb{P} \left( \omega \right)$.
    Now, we simply have to show that $M_f$ is a Markov kernel, which then proves the statement via the information monotonicity of f-divergences. \\
    For $A \in \mathcal{F}_f$ we have
    \begin{equation}
    \begin{split}
        M_f\mathbb{P} \left( A \right) & = \int_\Omega M_f \left( \omega, A \right) \mathrm{d} \mathbb{P} \left( \omega \right) \\
        & = \int_\Omega \mathbbm{1}_{\left\{ \omega \in \Omega \mid f \left( \omega \right) \in A \right\}} \left( \omega \right) \mathrm{d} \mathbb{P} \left( \omega \right) \\
        & = \int_{\left\{ \omega \in \Omega \mid f \left( \omega \right) \in A \right\}} \mathrm{d} \mathbb{P} \\
        & = \mathbb{P} \left( \left\{ \omega \in \Omega \mid f \left( \omega \right) \in A \right\} \right).
    \end{split}
    \end{equation}
    The last line shows that $M_f \mathbb{P}$ maps into a distribution space, which fulfills the definition of a Markov kernel as given in \citep{garcia2012divergences}. \\
    Now, using the information monotonicity of f-divergences, we get for $i \in \left\{2, \dots, l \right\}$
    \begin{equation}
    \begin{split}
        I_{F^Y} \left( P^{{i}}_1, \dots, P^{{i}}_k \parallel P^{{i}} \right) & = I_{F^Y} \left( M_{g_i} P^{{i-1}}_1, \dots, M_{g_i} P^{{i-1}}_k \parallel M_{g_i} P^{{i-1}} \right) \\
        & \leq I_{F^Y} \left( P^{{i-1}}_1, \dots, P^{{i-1}}_k \parallel P^{{i-1}} \right),
    \end{split}
    \end{equation}
    which proves the inequality chain.
    It is upper bounded by the statistical information since
    \begin{equation}
    \begin{split}
        & I_{F^Y} \left( P^{{1}}_1, \dots, P^{{1}}_k \parallel P^{{1}} \right) \\
        & = I_{F^Y} \left( M_{g_1} \mathbb{P} \left( X \mid Y=1 \right), \dots, M_{g_1} \mathbb{P} \left( X \mid Y=k \right) \parallel M_{g_1} \mathbb{P} \left( X \right) \right) \\
        & \leq I_{F^Y} \left( \mathbb{P} \left( X \mid Y=1 \right), \dots, \mathbb{P} \left( X \mid Y=k \right) \parallel \mathbb{P} \left( X \right) \right) \\
        & = \int_{\mathcal{X}} F \left(\mathbb{E} \left[ \Vec{Y}_1 \right] \frac{\mathrm{d}\mathbb{P} \left( X \mid Y = 1 \right)}{\mathrm{d} \mathbb{P} \left( X \right)}, \dots, \mathbb{E} \left[ \Vec{Y}_k \right] \frac{\mathrm{d}\mathbb{P} \left( X \mid Y = k \right)}{\mathrm{d} \mathbb{P} \left( X \right)} \right) - F \left( \mathbb{E} \left[ \Vec{Y} \right] \right) \mathrm{d} \mathbb{P} \left( X \right) \\
        & = \mathbb{E} \left[ F \left( \mathbb{E} \left[ \Vec{Y} \mid X \right] \right) \right] - F \left( \mathbb{E} \left[ \Vec{Y} \right] \right) \\
        & = \operatorname{SI}_F \left(X; Y \right).
    \end{split}
    \end{equation}
\end{proof}

\section{Class-wise calibration induced by One-vs-Rest risk}
\label{appendix:classwise_calibration_one_v_rest}

In this section, we derive class-wise calibration errors from One-vs-Rest risk minimization.
We do so by decomposing the One-vs-Rest risk into calibration and sharpness terms analogous to the standard risk minimization case.
This is a novel contribution towards better understanding of class-wise calibration errors.
Specifically, this suggests to use class-wise calibration in predictive scenarios when the multi-class prediction consists of probabilities, which do not sum up to one.

For a binary loss $L \colon \left[ 0, 1 \right] \times \left\{0, 1\right\} \to \mathbb{R}$,
we define the risk of class $i$ vs the rest as

\begin{equation}
    \mathcal{R}_i \left( g \right) \coloneqq \mathbb{E} \left[ L \left( g \left( X \right), \mathbbm{1}\left\{ Y = i \right\} \right) \right],
\end{equation}

where $g \colon \mathcal{X} \to \left[0, 1 \right]$ is a binary classifier and $Y$ takes values in $\left\{1, \dots, k \right\}$.

In the following, we assume $-L$ is a proper score, i.e. it is minimized by the Bayes classifier.

Analogous, the negative Bayes risk for a $q \in \left[0, 1 \right]$ is given by

\begin{equation}
    F \left( q \right) \coloneqq - \inf_{p \in \left[0, 1 \right]} \mathbb{E}_{Y \sim q} \left[ L \left( p, Y \right) \right].
\end{equation}

Like in the canonical case, $F$ is convex as long as $-L$ is a proper score.
Consequently, $D_F$ is a Bregman divergence.
The Brier score and the squared Euclidean distance are recovered by $F \left( q \right) = q^2 + \left( 1 - q \right)^2$.
The negative log likelihood and the Kullback--Leibler divergence are given by the case $F \left( q \right) = q \log q + \left( 1 - q \right) \log \left( 1 - q \right)$.
As a novel contribution, we derive class-wise calibration errors from the mean of the class-wise One-vs-Rest risk of binary classifiers $g_1, \dots, g_k$ as

\begin{equation}
    \frac{1}{k} \sum_{i=1}^k \mathcal{R}_i \left( g_i \right) = \frac{1}{k} \sum_{i=1}^k - F \left( \mathbb{P} \left( Y = i \right) \right) + \operatorname{CWCE}_F \left( g_1, \dots, g_k \right) - \operatorname{SHARP}_F \left( g_1, \dots, g_k \right)
\end{equation}

where we have a class-wise calibration error $\operatorname{CWCE}_F \left( g_1, \dots, g_k \right) \coloneqq \frac{1}{k} 
\sum_{i=1}^k \mathbb{E} \left[ D_F \left( \mathbb{P}\left( Y = i \mid g_i \left( X \right) \right), g_i \left( X \right) \right) \right]$ and a class-wise sharpness $\operatorname{SHARP}_F \left( g_1, \dots, g_k \right) \coloneqq \frac{1}{k} \sum_{i=1}^k \mathbb{E} \left[ D_F \left( \mathbb{P}\left( Y = i \mid g_i \left( X \right), \mathbb{P}\left( Y = i \right) \right) \right) \right]$.
Analogous to the canonical case, we have $\operatorname{CWCE}_F \left( g_1, \dots, g_k \right) = \operatorname{CWCE}_2^2 \left( g \right)$ for $F \left( q \right) = q^2$ and $g \left( x \right) \coloneqq \left(g_1 \left( x \right), \dots, g_k \left( x \right) \right)^\intercal$.
Surprisingly, the associated negative Bayes risk is not symmetric unlike other common cases \citep{gneitingscores}.
Note that the factor $\frac{1}{k}$ systematically decreases the class-wise calibration error with growing $k$ if the class-wise predictions are normalized to a probability vector \citep{gruber2022better}.

\begin{proof}
We first show that the decomposition holds.
Note that we implied $F$ is differentiable, but the decomposition still holds under general conditions by replacing Bregman divergences with score divergences \citep{gneitingscores}.
Since by assumption $L$ is a negative proper score, we have $L \left( p, y \right) = - F \left( p \right) - F^\prime \left( p \right) \left( y - p \right)$ for $p \in \left[0, 1 \right]$ and $y \in \left\{ 0, 1 \right\}$ (c.f. Schervish representation \citep{gneitingscores}). \\
Further, note that for $i \in \left\{1, \dots, k \right\}$ we have
\begin{equation}
\begin{split}
    & \mathcal{R}_i \left( g_i \right) \\
    \overset{\text{def}}&{=} \mathbb{E} \left[ L \left( g_i \left( X \right), \mathbbm{1}\left\{ Y = i \right\} \right) \right] \\
    & = \mathbb{E} \left[ - F \left( g_i \left( X \right) \right) - F^\prime \left( g_i \left( X \right) \right) \left( \mathbbm{1}\left\{ Y = i \right\} - g_i \left( X \right) \right) \right] \\
    \overset{\text{LOTUS}}&{=} \mathbb{E} \left[ - F \left( g_i \left( X \right) \right) - F^\prime \left( g_i \left( X \right) \right) \left( \mathbb{P} \left( Y = i \mid g_i \left( X \right) \right) - g_i \left( X \right) \right) \right] \\
    & = \mathbb{E} \left[ D_F \left( \mathbb{P} \left( Y = i \mid g_i \left( X \right) \right), g_i \left( X \right) \right) \right] - \mathbb{E} \left[ F \left( \mathbb{E} \left( Y = i \mid g_i \left( X \right) \right) \right) \right] \\
    & = \mathbb{E} \left[ D_F \left( \mathbb{P} \left( Y = i \mid g_i \left( X \right) \right), g_i \left( X \right) \right) \right] - \mathbb{E} \left[ D_F \left( \mathbb{P} \left( Y = i \mid g_i \left( X \right) \right), \mathbb{P} \left( Y = i \right) \right) \right] \\
    & \quad \quad - \mathbb{E} \left[ F \left( \mathbb{P} \left( Y = i \mid g_i \left( X \right) \right) \right) \right],
\end{split}
\end{equation}

where we used the law of the unconscious statistician (LOTUS) and the definition of Bregman divergences. \\
Now, we use this result to get

\begin{equation}
\begin{split}
    \frac{1}{k} \sum_{i=1}^k \mathcal{R}_i \left( g_i \right) & = \frac{1}{k} \sum_{i=1}^k - F \left( \mathbb{P} \left( Y = i \right) \right) \\
    & \quad \quad + \frac{1}{k} \sum_{i=1}^k \mathbb{E} \left[ D_F \left( \mathbb{P} \left( Y = i \mid g_i \left( X \right) \right), g_i \left( X \right) \right) \right] \\
    & \quad \quad - \frac{1}{k} \sum_{i=1}^k \mathbb{E} \left[ D_F \left( \mathbb{P} \left( Y = i \mid g_i \left( X \right) \right), \mathbb{P} \left( Y = i \right) \right) \right] \\
    \overset{\text{def}}&{=} \frac{1}{k} \sum_{i=1}^k - F \left( \mathbb{P} \left( Y = i \right) \right) + \operatorname{CWCE}_F \left( g_1, \dots, g_k \right) - \operatorname{SHARP}_F \left( g_1, \dots, g_k \right).
\end{split}
\end{equation}

Last, we show $\operatorname{CWCE}_F \left( g_1, \dots, g_k \right) = \operatorname{CWCE}_2^2 \left( g \right)$ for $F \left( q \right) = q^2$ and $g \left( x \right) \coloneqq \left(g_1 \left( x \right), \dots, g_k \left( x \right) \right)^\intercal$.
Since $D_F \left(x, y \right) = \left( x - y \right)^2$, we have

\begin{equation}
\begin{split}
    \operatorname{CWCE}_F \left( g_1, \dots, g_k \right) \overset{\text{def}}&{=} \frac{1}{k} \sum_{i=1}^k \mathbb{E} \left[ D_F \left( \mathbb{P} \left( Y = i \mid g_i \left( X \right) \right), g_i \left( X \right) \right) \right] \\
    & = \frac{1}{k} \sum_{i=1}^k \mathbb{E} \left[ \left( \mathbb{P} \left( Y = i \mid g_i \left( X \right) \right) - g_i \left( X \right) \right)^2 \right] \\
    \overset{\text{def}}&{=} \operatorname{CWCE}_2^2 \left( g \right).
\end{split}
\end{equation}
\end{proof}

\section{Proper calibration error estimators via KDE}
\label{appendix:bregman_divergence}

\subsection{The Dirichlet kernel in estimating $\mathbb{E}[Y\mid g(x)]$}

The Dirichlet kernel is defined as: 
\begin{equation}
    k_{Dir}(g(x_h),g(x_j)) = \frac{\Gamma(\sum_{k=1}^K \alpha_{jk})}{\prod_{k=1}^K \Gamma(\alpha_{jk})} \prod_{k=1}^K g(x_h)_{k}^{\alpha_{jk}-1}
\end{equation}
with $\alpha_{j} = \frac{g(x_j)}{\gamma} + 1$, $\gamma>0$ being a bandwidth parameter \citep{ouimet2020,popordanoska2022}.
We note that popular libraries such as PyTorch provide only $\log\Gamma$\footnote{\url{https://pytorch.org/docs/stable/generated/torch.lgamma.html}} and not $\Gamma$ directly \citep{NEURIPS2019_bdbca288}.  It will therefore be useful to consider
\begin{align}
    \log(k(g(x_h),g(x_j))) =&  \log \frac{\Gamma(\sum_{k=1}^K \alpha_{jk})}{\prod_{k=1}^K \Gamma(\alpha_{jk})} \prod_{k=1}^K g(x_h)_{k}^{\alpha_{jk}-1} \\
     =&  \log\Gamma\left(\sum_{k=1}^K \alpha_{jk}\right) - \sum_{k=1}^K \log\Gamma\left(\alpha_{jk}\right) + \sum_{k=1}^K \left(\alpha_{jk}-1\right)\log g(x_h)_{k} \\
     =& \log\Gamma\left(K+ \sum_{k=1}^K \frac{g(x_j)_k}{\gamma} \right) - \sum_{k=1}^K \log\Gamma\left(\frac{g(x_j)_k}{\gamma
     } + 1 \right) + \frac{1}{\gamma} \langle g(x_j), \log(g(x_h))\rangle     
\end{align}
and we can further apply the log to the softmax function in the last $\log(g(x_h))$ inside the inner product.

We subsequently focus on terms of the form
\begin{align}
    \log \left(\sum_{j\neq h} k(g(x_h),g(x_j))\right) =&  \operatorname{LogSumExp}_{j\neq h}\left( \log(k(g(x_h),g(x_j)))\right) .
\end{align}
  Computation of terms of the form $\log \left(\sum_{j\neq h} k(g(x_h),g(x_j))y_j\right)$ are essentially the same, but the LogSumExp operation should only be performed over indices where $y_{jk}\neq 0$.

\subsection{Bregman derivation of the $L_2$ calibration error estimator}
\label{sec:BregmalL2CE}

For the Bregman formulation of squared $L_2$ error, we have $F(x) = \|x\|_2^2$, and the r.h.s.\ of Equation~\eqref{eq:BregmanCalibrationError} becomes
\begin{align} \nonumber
\frac{1}{n} \sum_{h=1}^n \Bigg(  \left\| \frac{\sum_{j\neq h} k(g(x_h),g(x_j))y_j}{\sum_{j\neq h} k(g(x_h),g(x_j))} \right\|_2^2 + \|g(x_h)\|_2^2 -  2 \left\langle g(x_h) , \frac{\sum_{j\neq h} k(g(x_h),g(x_j))y_j}{\sum_{j\neq h} k(g(x_h),g(x_j))}   \right\rangle \Bigg) \\
=\frac{1}{n} \sum_{h=1}^n  \left\| \frac{\sum_{j\neq h} k(g(x_h),g(x_j))y_j}{\sum_{j\neq h} k(g(x_h),g(x_j))} - g(x_h) \right\|_2^2  .
\end{align}
We see that this recovers exactly the $L_2$ special case of the estimator given by \cite[Equation(9)]{popordanoska2022}.  That paper shows that the resulting estimator is consistent, and has a convergence of $\mathcal{O}(n^{-1/2})$ with a bias that converges as $\mathcal{O}(n^{-1})$.

\subsection{Bregman derivation of the KL calibration error estimator}

Recall from above that the Bregman KL divergence is generated by $F(p) = \langle p, \log(p)\rangle$, where the $\log$ operation is applied element-wise.

The Bregman formulation of KL calibration error is
\begin{align}
\frac{1}{n}\sum_{h=1}^n \Bigg(
\left\langle \frac{\sum_{j\neq h} k(g(x_h),g(x_j))y_j}{\sum_{j\neq h} k(g(x_h),g(x_j))} , \log \left( \frac{\sum_{j\neq h} k(g(x_h),g(x_j))y_j}{\sum_{j\neq h} k(g(x_h),g(x_j))} \right) \right\rangle - \langle g(x_h), \log(g(x_h)) \rangle - \nonumber \\
\left\langle 
\log(g(x_h)) + e
, \frac{\sum_{j\neq h} k(g(x_h),g(x_j))y_j}{\sum_{j\neq h} k(g(x_h),g(x_j))}  - g(x_h)  \right\rangle
\Bigg) \nonumber\\
=\frac{1}{n}\sum_{h=1}^n 
\left\langle \frac{\sum_{j\neq h} k(g(x_h),g(x_j))y_j}{\sum_{j\neq h} k(g(x_h),g(x_j))} , \log \left( \frac{\sum_{j\neq h} k(g(x_h),g(x_j))y_j}{\sum_{j\neq h} k(g(x_h),g(x_j)) g(x_h)} \right) 
\right\rangle 
\end{align}
where $e$ is a vector of all ones and division of two vectors is assumed to be element-wise.  The result is an estimator identical to Equation~\eqref{eq:KLcalibrationError}. 

\subsection{General Sharpness-Calibration error decompositions of expectations of Bregman divergences}

In general, we can define a statistical risk measure based on a Bregman divergence as
\begin{align}
    \mathbb{E}_{(X,Y)\sim p} [ D_F(Y, g(X)) ] = \mathbb{E}_{(X,Y)} [ F(Y) - F(g(X)) - \langle \nabla F(g(X)),Y - g(X) \rangle ] .
\end{align}
If we subtract out the associated Bregman calibration error (cf.\ Equation~\eqref{eq:BregmanCalibrationError}), we have
\begin{align}
    \mathbb{E}[ D_F(Y, g(X)) ] - \operatorname{CE}_{F}(g) =& \mathbb{E}_{(X,Y)} [ D_F(Y, g(X)) ]  - \mathbb{E}_X \left[D_{F}(\mathbb{E}[Y \mid g(X)] , g(X) ) \right] \\
    =& \mathbb{E} [ F(Y) - F(g(X)) - \langle \nabla F(g(X)),Y - g(X) \rangle] 
    \\ \nonumber
    & - \mathbb{E} [
    F( \mathbb{E}[Y \mid g(X)]) - F(g(X)) - \langle \nabla F(g(X)), \mathbb{E}[Y \mid g(X)] - g(X) \rangle 
     ] \\
     =& \mathbb{E} [ F(Y)  - 
    F( \mathbb{E}[Y \mid g(X)])] - \mathbb{E}[\langle \nabla F(g(X)),Y - \mathbb{E}[Y \mid g(X)]\rangle 
     ]  .
\end{align}
It is a well known property of conditional expectation that $\mathbb{E}[f(Z)\cdot Y\mid Z] = f(Z) \mathbb{E}[Y\mid Z]$, which yields
\begin{align}
    \mathbb{E}[\langle \nabla F(g(X)),Y - \mathbb{E}[Y \mid g(X)]\rangle 
     ] =& \mathbb{E}[\langle \nabla F(g(X)),Y \rangle] - \underbrace{\mathbb{E}[\mathbb{E}[\langle \nabla F(g(X)), Y\rangle \mid g(X)] 
     ]}_{=\mathbb{E}[\langle \nabla F(g(X)),Y \rangle] \text{ by law of total expectation}} = 0,
\end{align}
and our equation simplifies to
\begin{align}
\label{eq:decomposition_risk_ce_refinement}
    \mathbb{E}[ D_F(Y, g(X)) ]  - \operatorname{CE}_{F}(g) = \mathbb{E} [ F(Y)  - 
    F( \mathbb{E}[Y \mid g(X)])] .
\end{align}

\subsubsection{Recovery of $L_2$ refinement}

Setting $F(p) = \|p\|^2$, the risk becomes the Brier score and we obtain
\begin{align}
 \mathbb{E} [ \|Y\|^2  - 
    \| \mathbb{E}[Y \mid g(X)] \|^2] 
     =& 1  - 
      \mathbb{E}_{X} [\|\mathbb{E}[Y \mid g(X)] \|^2]
\end{align}
\cite{popordanoska2022} show the $L_2$ refinement to be
 $\mathbb{E}[(1-\mathbb{E}[Y\mid g(X))]\mathbb{E}[Y\mid g(X)]]=\mathbb{E}[\mathbb{E}[Y\mid g(X)] - \mathbb{E}[Y\mid g(X)]^2] = \mathbb{E}[Y] - \mathbb{E}[\mathbb{E}[Y\mid g(X)]^2]$.  This is in fact the first term of the above if we expand the norm as a sum over elements and split into expectations of terms from each dimension of $Y$.

\subsubsection{Recovery of KL refinement}

Setting $F(p) = \langle p, \log(p)\rangle$, the risk corresponds with cross-entropy loss, and we obtain
\begin{align}
    \mathbb{E} [\underbrace{\langle Y, \log(Y)\rangle}_{=0} - \langle \mathbb{E}[Y\mid g(X)],\log(\mathbb{E}[Y\mid g(X)]\rangle] 
    =&- \mathbb{E}[\langle \mathbb{E}[Y\mid g(X)],\log(\mathbb{E}[Y\mid g(X)]\rangle] \\
    =& \mathbb{E}[H(\mathbb{E}[Y\mid g(X)])] ,
\end{align}
where $H$ denotes entropy, we interpret the first inner product involving $Y$ using $\lim_{Y\searrow 0} Y \log Y = 0$ as it is otherwise undefined, and we additionally assume categorical labels $Y$ without uncertainty.

\section{Experiments}
\label{appendix:experiments}

In this section, we first provide a detailed description of the empirical setup. Subsequently, we further investigate the bias convergence of our estimator, both on simulated and real-world datasets. 
Then, we show a table evaluating accuracy, NLL and Brier score on CIFAR 10/100 before and after calibration with isotonic regression and temperature scaling.
Finally, we present additional results for several use-cases of the estimator: monitoring model training, model selection, and assessing calibration error.

\subsection{Empirical setup}

\paragraph{Datasets}
The experiments in the main text rely on two widely used benchmark datasets, CIFAR-10/100 \citep{krizhevsky2009learning}, which consist of $32 \times 32$ natural images divided into 10 and 100 classes, respectively. We split the data into train/validation/test sets of 45000/5000/10000.

\paragraph{Models} 
We trained PreResNet20, PreResNet56, PreResNet110, PreResNet164~\citep{he2016identity}, VGG16 (with BatchNorm)~\citep{simonyan2014very} and WideResNet28x10~\citep{zagoruyko2016wide} for 250 epochs with Stochastic Gradient Descent optimizer using PyTorch~\citep{paszke2017automatic}. The learning rate was reduced by a factor of 10 at $150^{th}$ and $225^{th}$ epochs.
The WideResNet and the PreResNet models were trained with the learning rate of $1\mathrm{e}{-1}$, batch size of $128$, weighted decay (WD) of $1\mathrm{e}{-10}$ and Nesterov's momentum of 0.9. During the first epoch, we warmed up the training with the learning rate of $0.01$. The VGG model was trained with the learning rate of $5\mathrm{e}{-2}$, WD of $5\mathrm{e}{-5}$, batch size of $128$, and momentum of $0.05$. The training was carried out with NVIDIA V100 GPU.

\subsection{Convergence of bias}

\begin{figure}[ht]
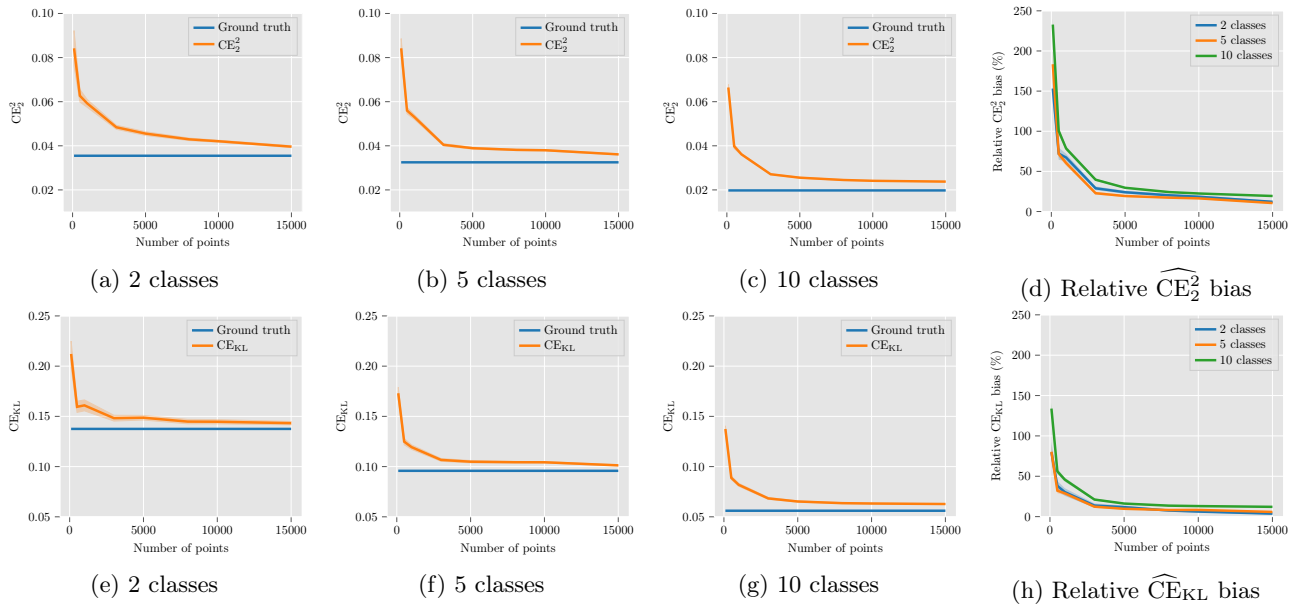

    \centering
    \subfloat[2 classes]{
    \resizebox{0.24\textwidth}{!}{\input{figs/bias_convergence_synthetic_l2_2classes}}
    }
    \subfloat[5 classes]{
    \resizebox{0.24\textwidth}{!}{\input{figs/bias_convergence_synthetic_l2_5classes}}
    }
    \subfloat[10 classes]{
    \resizebox{0.24\textwidth}{!}
    {\input{figs/bias_convergence_synthetic_l2_10classes}}
    }
    \subfloat[Relative $\widehat{\operatorname{CE}_{2}^{2}}$ bias]{
    \resizebox{0.24\textwidth}{!}
    {\input{figs/bias_convergence_synthetic_l2_rel_bias}}
    }
    \hfill
    \subfloat[2 classes]{
    \resizebox{0.24\textwidth}{!}{\input{figs/bias_convergence_synthetic_kl_2classes}}
    }
    \subfloat[5 classes]{
    \resizebox{0.24\textwidth}{!}{\input{figs/bias_convergence_synthetic_kl_5classes}}
    }
    \subfloat[10 classes]{
    \resizebox{0.24\textwidth}{!}
    {\input{figs/bias_convergence_synthetic_kl_10classes}}
    }
    \subfloat[Relative $\widehat{\operatorname{CE}}_{\mathrm{KL}}$ bias]{
    \resizebox{0.24\textwidth}{!}
    {\input{figs/bias_convergence_synthetic_kl_rel_bias}}
    }
    \caption{Calibration error and relative bias (\%) on simulated data for different number of classes. Each plot shows the estimate as a function of the sample size. The top row evaluates $\widehat{\operatorname{CE}_{2}^{2}}$, while the bottom row $\widehat{\operatorname{CE}}_{\mathrm{KL}}$.}
    \label{fig:bias_convergence_synthetic_appendix}
\end{figure}

In addition to our empirical analysis for the convergence of bias on simulated data in the main part, here we present the calibration error and the relative bias, computed as $\frac{\widehat{\operatorname{CE}} - \operatorname{CE}}{\operatorname{CE}}*100$, with $\widehat{\operatorname{CE}}$ the estimated and $\operatorname{CE}$ the ground truth calibration error. The averaged results across 20 iterations of sampling new points for the estimation, together with the standard errors, are shown in Figure~\ref{fig:bias_convergence_synthetic_appendix}.

\begin{figure}[ht]
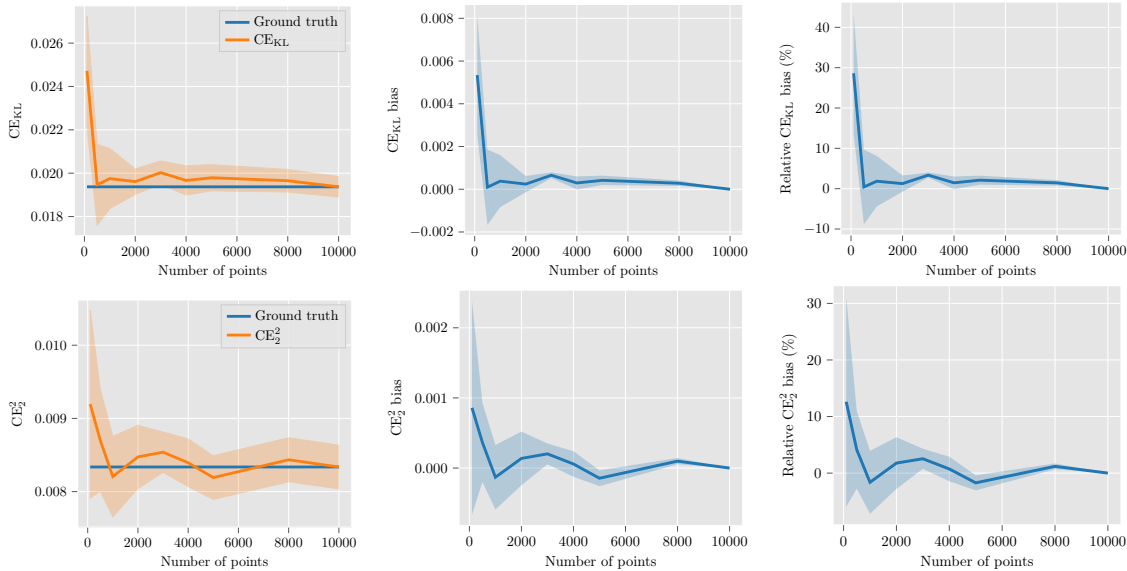

    \centering
    \subfloat{
    \resizebox{0.28\textwidth}{!}{\input{figs/bias_convergence_cifar10_PreResNet56_kl}}
    }
    \subfloat{
    \resizebox{0.29\textwidth}{!}{\input{figs/bias_convergence_cifar10_PreResNet56_bias_kl}}
    }
    \subfloat{
    \resizebox{0.28\textwidth}{!}
    {\input{figs/bias_convergence_cifar10_PreResNet56_rel_bias_kl}}
    }
    \hfill
    \subfloat{
    \resizebox{0.28\textwidth}{!}{\input{figs/bias_convergence_cifar10_PreResNet56_l2}}
    }
    \subfloat{
    \resizebox{0.29\textwidth}{!}{\input{figs/bias_convergence_cifar10_PreResNet56_bias_l2}}
    }
    \subfloat{
    \resizebox{0.28\textwidth}{!}
    {\input{figs/bias_convergence_cifar10_PreResNet56_rel_bias_l2}}
    }
    \caption{Calibration error, bias, and relative bias (\%) evaluated on the test set of CIFAR 10, using predictions of trained PreResNet56 architectures with four different seeds. The ground truth is obtained from the whole test set.}
    \label{fig:bias_convergence_cifar_appendix}
\end{figure}

Similarly, in Figure~\ref{fig:bias_convergence_cifar_appendix} we evaluate the calibration error, bias and relative bias on the test set of CIFAR-10, as a function of the number of points used for the estimation. The ground truth is calculated from the whole test set (10000 images). The bandwidth of the Dirichlet kernel is set to 0.02. The results reveal that the estimator achieves values that closely align with the ground truth, even with as few as a (couple of) hundred points.

\subsection{Post-hoc calibration}
In addition to the experiment in the main text, where we evaluate $\widehat{\operatorname{CE}_{2}^{2}}$ and $\widehat{\operatorname{CE}}_{\mathrm{KL}}$ before and after calibration, in Table~\ref{tab:calibration_methods_perf} we show accuracy, NLL and Brier score of various network architectures on CIFAR-10/100. 

\begin{table*}[ht]
    \centering
    \caption{Performance evaluation (Acc $\times 100$, NLL $\times 100$ and Brier score $\times 100$) of various network architectures on CIFAR-10/100 with no calibration, and recalibration with Isotonic Regression and Temperature Scaling.}
    \label{tab:calibration_methods_perf}
\resizebox{1\textwidth}{!}{
    \begin{tabular}{ccccccccccc}
        \toprule
        & & \multicolumn{3}{c}{\textbf{No calibration}} & \multicolumn{3}{c}{\textbf{Isotonic Regression}} & \multicolumn{3}{c}{\textbf{Temperature Scaling}} \\
        Dataset & Model & Acc & NLL & Brier & Acc & NLL & Brier & Acc & NLL & Brier \\
        \midrule
        \multirow{6}{*}{CIFAR-10} 
        & PreResNet20 & $91.95_{\pm 0.05}$ & $32.65_{\pm 0.45}$ & $12.75_{\pm 0.08}$ & $91.94_{\pm 0.07}$ & $26.92_{\pm 0.09}$ & $11.99_{\pm 0.07}$ & $91.95_{\pm 0.05}$ & $24.11_{\pm 0.14}$ & $11.74_{\pm 0.07}$ \\ 
        & PreResNet56 & $94.38_{\pm 0.13}$ & $22.46_{\pm 0.25}$ & $9.03_{\pm 0.18}$ & $94.34_{\pm 0.13}$ & $19.58_{\pm 0.19}$ & $8.61_{\pm 0.14}$ & $94.38_{\pm 0.13}$ & $17.48_{\pm 0.17}$ & $8.41_{\pm 0.13}$ \\ 
        & PreResNet110 & $94.86_{\pm 0.04}$ & $20.42_{\pm 0.18}$ & $8.24_{\pm 0.06}$ & $94.83_{\pm 0.05}$ & $18.06_{\pm 0.34}$ & $7.91_{\pm 0.04}$ & $94.86_{\pm 0.04}$ & $16.11_{\pm 0.07}$ & $7.72_{\pm 0.06}$ \\ 
        & PreResNet164 & $95.24_{\pm 0.05}$ & $18.61_{\pm 0.29}$ & $7.58_{\pm 0.06}$ & $95.14_{\pm 0.06}$ & $17.39_{\pm 0.33}$ & $7.33_{\pm 0.07}$ & $95.24_{\pm 0.05}$ & $14.96_{\pm 0.17}$ & $7.13_{\pm 0.06}$ \\ 
        & VGG16BN & $93.26_{\pm 0.04}$ & $33.76_{\pm 0.29}$ & $11.61_{\pm 0.10}$ & $93.23_{\pm 0.04}$ & $25.15_{\pm 0.33}$ & $10.42_{\pm 0.08}$ & $93.26_{\pm 0.04}$ & $24.55_{\pm 0.16}$ & $10.48_{\pm 0.09}$ \\ 
        & WideResNet28x10 & $95.54_{\pm 0.05}$ & $15.69_{\pm 0.16}$ & $7.00_{\pm 0.07}$ & $95.53_{\pm 0.04}$ & $16.49_{\pm 0.44}$ & $6.86_{\pm 0.08}$ & $95.54_{\pm 0.05}$ & $14.50_{\pm 0.17}$ & $6.80_{\pm 0.09}$ \\ 
        \midrule
        \multirow{6}{*}{CIFAR-100} 
        & PreResNet20 & $68.01_{\pm 0.15}$ & $121.48_{\pm 1.12}$ & $44.38_{\pm 0.15}$ & $67.58_{\pm 0.13}$ & $134.53_{\pm 1.92}$ & $44.60_{\pm 0.07}$ & $68.01_{\pm 0.15}$ & $113.17_{\pm 0.30}$ & $42.95_{\pm 0.09}$ \\ 
        & PreResNet56 & $74.38_{\pm 0.10}$ & $112.80_{\pm 2.88}$ & $38.24_{\pm 0.43}$ & $74.00_{\pm 0.09}$ & $115.75_{\pm 1.44}$ & $36.88_{\pm 0.17}$ & $74.38_{\pm 0.10}$ & $93.10_{\pm 1.02}$ & $35.41_{\pm 0.21}$ \\ 
        & PreResNet110 & $75.62_{\pm 0.14}$ & $106.74_{\pm 0.53}$ & $36.44_{\pm 0.17}$ & $75.26_{\pm 0.09}$ & $108.38_{\pm 1.00}$ & $35.25_{\pm 0.15}$ & $75.62_{\pm 0.14}$ & $88.82_{\pm 0.45}$ & $33.83_{\pm 0.15}$ \\ 
        & PreResNet164 & $76.54_{\pm 0.05}$ & $103.19_{\pm 0.55}$ & $35.12_{\pm 0.13}$ & $76.11_{\pm 0.10}$ & $108.24_{\pm 0.85}$ & $34.18_{\pm 0.12}$ & $76.54_{\pm 0.05}$ & $86.30_{\pm 0.54}$ & $32.68_{\pm 0.13}$ \\ 
        & VGG16BN & $71.36_{\pm 0.08}$ & $167.78_{\pm 0.98}$ & $46.45_{\pm 0.16}$ & $71.08_{\pm 0.18}$ & $137.77_{\pm 0.91}$ & $40.96_{\pm 0.14}$ & $71.36_{\pm 0.08}$ & $120.41_{\pm 0.51}$ & $39.77_{\pm 0.13}$ \\ 
        & WideResNet28x10 & $79.56_{\pm 0.22}$ & $84.17_{\pm 0.80}$ & $29.93_{\pm 0.31}$ & $79.23_{\pm 0.21}$ & $99.28_{\pm 1.49}$ & $29.92_{\pm 0.28}$ & $79.56_{\pm 0.22}$ & $81.83_{\pm 0.68}$ & $29.44_{\pm 0.29}$ \\ 
        \bottomrule
    \end{tabular}
    }
\end{table*} 

\subsection{Additional experiments}
\paragraph{Monitoring calibration error during training}

Monitoring accuracy and loss during the training of a deep neural network is essential for various reasons, including performance evaluation and model selection. 
We argue that monitoring calibration error and sharpness/refinement, in addition to the standard metrics, provides additional insights into the reliability and performance of the model.

\begin{figure*}[ht]
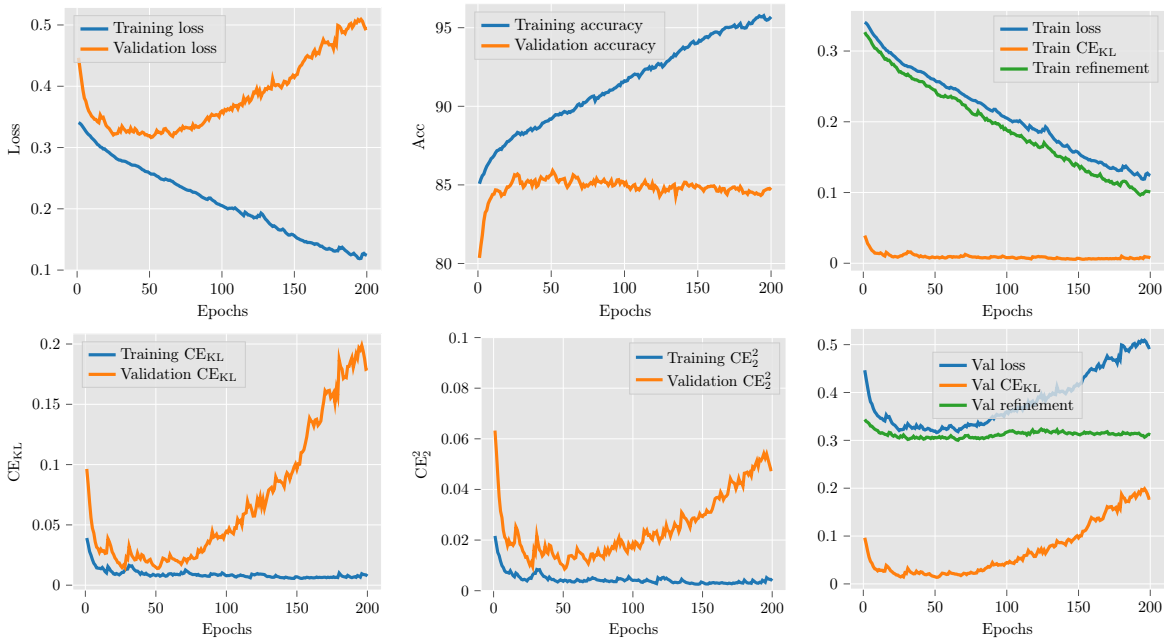

    \centering
    \subfloat{
    \resizebox{0.3\textwidth}{!}{\input{figs/breast_cancer_kaggle_VGG16BN_loss_}}
    \label{subfig:cancer_loss}
    }
    \subfloat{
    \resizebox{0.3\textwidth}{!}{\input{figs/breast_cancer_kaggle_VGG16BN_acc_}}
    \label{subfig:cancer_acc}
    }
    \subfloat{
    \resizebox{0.28\textwidth}{!}
    {\input{figs/breast_cancer_kaggle_VGG16BN_train_kl}}
    \label{subfig:cancer_train}
    }
    \hfill    
    \subfloat{
    \resizebox{0.3\textwidth}{!}{\input{figs/breast_cancer_kaggle_VGG16BN_ce_kl}}
    \label{subfig:cancer_ce_kl}
    }
    \subfloat{
    \resizebox{0.3\textwidth}{!}{\input{figs/breast_cancer_kaggle_VGG16BN_ce_l2}}
    \label{subfig:cancer_ce_l2}
    }
    \subfloat{
    \resizebox{0.28\textwidth}{!}
    {\input{figs/breast_cancer_kaggle_VGG16BN_val_kl}}
    \label{subfig:cancer_val}
    }
    \caption{Training and validation trends monitoring loss, accuracy, calibration error and refinement. The calibration error is an effective tool for detecting overfitting. Observing the loss together with the induced calibration error and refinement offers unique insights for the performance of the model.}
    \label{fig:cancer_data}
\end{figure*}

We trained VGG16~\citep{simonyan2014very} on a binary classification task, using a publicly available dataset of breast histopathology images \citep{janowczyk2016cancer}. We used 10\% of the image patches from the dataset, ensuring that the original 70:30 ratio of negative to positive points is maintained. We apply a smoothing technique (exponential moving average) to improve clarity.
In Figure~\ref{fig:cancer_data} we show the training and validation metrics per epoch and we observe several trends. For instance, we notice that as the model overfits and validation loss increases, the refinement remains fairly flat, and the increase in validation loss is only due to the increasing calibration error. $\widehat{\operatorname{CE}}_{\mathrm{KL}}$ not only correctly uncovers an early stopping point (same as the loss), but it also offers a more refined view of the nature of overfitting: while the validation accuracy remains constant, the CE considerably increases. 
For these reasons, we advocate for incorporating the calibration metric induced by the given loss as part of the standard practice for monitoring the training process.

\paragraph{Model selection}

\begin{figure}
    \centering
    \resizebox{0.5\textwidth}{!}{
    \input{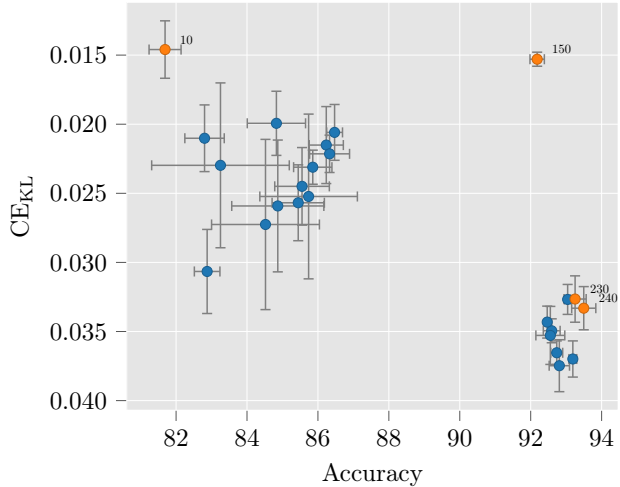}
    }
    \caption{Pareto front of VGG16 snapshots evaluated at every 10th epoch. The model is trained for a total of 250 epochs on CIFAR-10. The orange points Pareto-dominate the rest. The numbers represent the corresponding epoch. Note that the $y-$axis is inverted.}
    \label{fig:pareto_front}
\end{figure}

In practical settings, achieving high accuracy alone is often not sufficient for deploying machine learning models in decision-making pipelines. Obtaining a low calibration error becomes an equally important aspect in the evaluation of the model's performance. 
In such multi-objective optimization problems, the Pareto front is a useful tool to determine the set of optimal solutions. Figure~\ref{fig:pareto_front} shows the Pareto front of snapshots evaluated at every 10th epoch of training a VGG16 architecture on CIFAR-10 for a total of 250 epochs. Each point represents the mean and standard error across four seeds. The orange points are Pareto optimal (or Pareto efficient), meaning that no other snapshots can improve one objective without degrading the other. In other words, the orange points represent the best possible trade-off between accuracy and calibration error.

\paragraph{Assessing calibration error}

\begin{figure}[ht!]
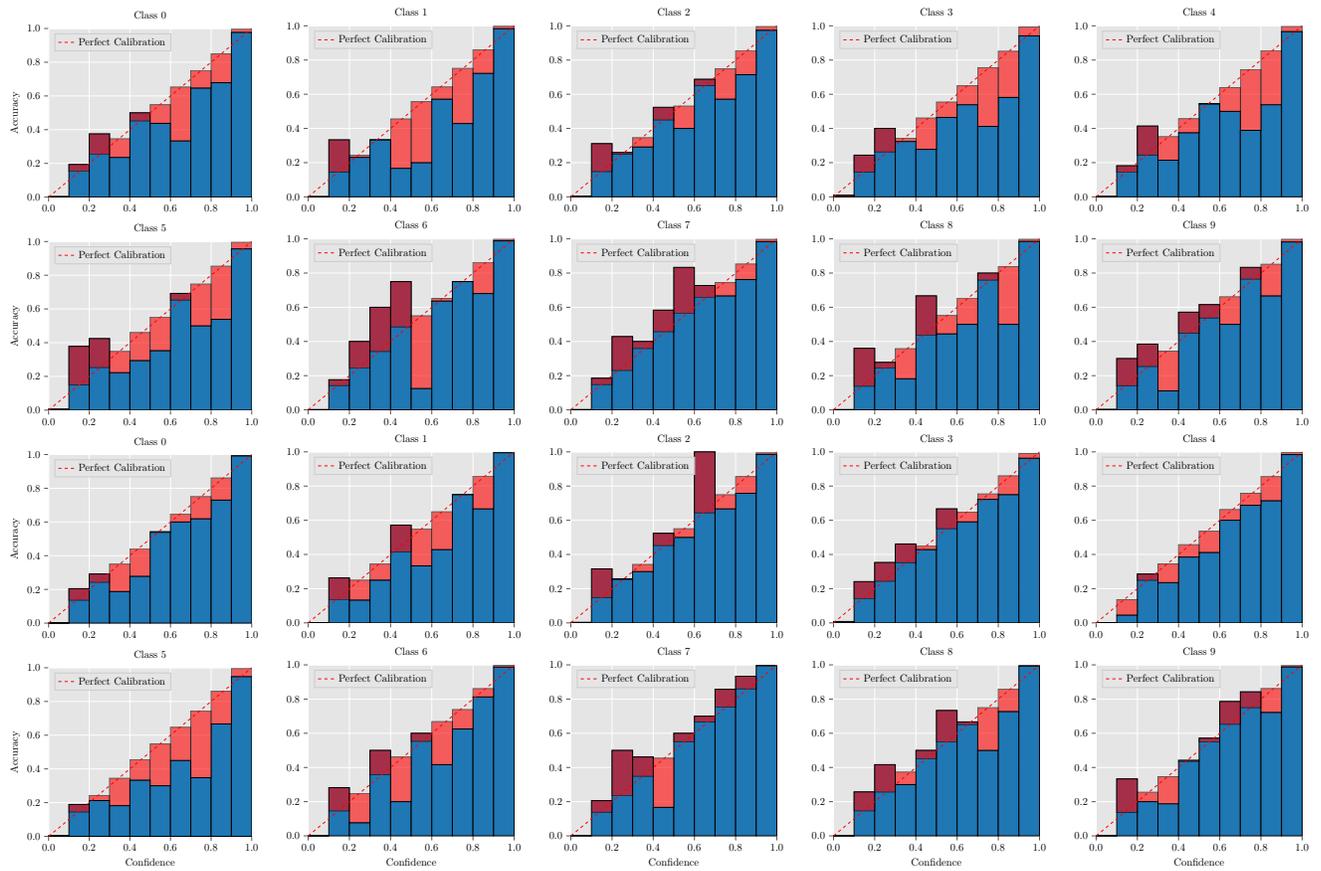

  \centering
    \subfloat{
        \resizebox{0.20\textwidth}{!}{\input{figs/reliability_diagram_cifar10_PreResNet56_10_class0}}
    }
    \subfloat{
        \resizebox{0.19\textwidth}{!}{\input{figs/reliability_diagram_cifar10_PreResNet56_10_class1}}
      }
    \subfloat{
        \resizebox{0.19\textwidth}{!}{\input{figs/reliability_diagram_cifar10_PreResNet56_10_class2}}
      }
      \subfloat{
        \resizebox{0.19\textwidth}{!}{\input{figs/reliability_diagram_cifar10_PreResNet56_10_class3}}
      }
      \subfloat{
        \resizebox{0.19\textwidth}{!}{\input{figs/reliability_diagram_cifar10_PreResNet56_10_class4}}
      } 
      \hfill
    \subfloat{
        \resizebox{0.20\textwidth}{!}{\input{figs/reliability_diagram_cifar10_PreResNet56_10_class5}}
    }
    \subfloat{
        \resizebox{0.19\textwidth}{!}{\input{figs/reliability_diagram_cifar10_PreResNet56_10_class6}}
      }
    \subfloat{
        \resizebox{0.19\textwidth}{!}{\input{figs/reliability_diagram_cifar10_PreResNet56_10_class7}}
      }
      \subfloat{
        \resizebox{0.19\textwidth}{!}{\input{figs/reliability_diagram_cifar10_PreResNet56_10_class8}}
      }
      \subfloat{
        \resizebox{0.19\textwidth}{!}{\input{figs/reliability_diagram_cifar10_PreResNet56_10_class9}}
      }
    \hfill
     \subfloat{
        \resizebox{0.20\textwidth}{!}{\input{figs/reliability_diagram_cifar10_WideResNet28x10_10_class0}}
    }
    \subfloat{
        \resizebox{0.19\textwidth}{!}{\input{figs/reliability_diagram_cifar10_WideResNet28x10_10_class1}}
      }
    \subfloat{
        \resizebox{0.19\textwidth}{!}{\input{figs/reliability_diagram_cifar10_WideResNet28x10_10_class2}}
      }
      \subfloat{
        \resizebox{0.19\textwidth}{!}{\input{figs/reliability_diagram_cifar10_WideResNet28x10_10_class3}}
      }
      \subfloat{
        \resizebox{0.19\textwidth}{!}{\input{figs/reliability_diagram_cifar10_WideResNet28x10_10_class4}}
      } 
      \hfill
    \subfloat{
        \resizebox{0.20\textwidth}{!}{\input{figs/reliability_diagram_cifar10_WideResNet28x10_10_class5}}
    }
    \subfloat{
        \resizebox{0.19\textwidth}{!}{\input{figs/reliability_diagram_cifar10_WideResNet28x10_10_class6}}
      }
    \subfloat{
        \resizebox{0.19\textwidth}{!}{\input{figs/reliability_diagram_cifar10_WideResNet28x10_10_class7}}
      }
      \subfloat{
        \resizebox{0.19\textwidth}{!}{\input{figs/reliability_diagram_cifar10_WideResNet28x10_10_class8}}
      }
      \subfloat{
        \resizebox{0.19\textwidth}{!}{\input{figs/reliability_diagram_cifar10_WideResNet28x10_10_class9}}
      }
    \hfill
    \caption{Reliability diagrams for each class of CIFAR-10 using PreResNet56 (top two rows) and WideResNet28x10 (bottom two rows).}
    \label{fig:reliability_plots_cifar10}
\end{figure}

In this part, we investigate the ranking performance of the class-wise versions of $\widehat{\operatorname{CE}}_{\mathrm{KL}}$ and $\widehat{\operatorname{CE}_{2}^{2}}$ via kernel density estimation, and $\operatorname{CE}_1$ with a binned estimator \citep{kull2019beyond, nixon2019measuring}.
The latter can be seen as the class-wise version of the commonly used ECE \citep{naeini2015, guo2017calibration}.
Specifically, we evaluate calibration error using the three estimators for each of the ten classes within the CIFAR-10 dataset. 
The similarity of their performance in terms of ranking the classes can be observed in Table~\ref{tab:classwise_metrics}, where the numbers in the brackets represent the ranking order. 
We used 15 bins with equal-width binning scheme for $\operatorname{CE}_1$, and the bandwidth for the KDE estimators was set to 0.02.
Additionally, Figure~\ref{fig:reliability_plots_cifar10} show the corresponding class-wise reliability diagrams for the models evaluated in the table. The blue bars represent the accuracy per bin. The red bars represent the gap of each bin to perfect calibration, i.e., the difference between accuracy and confidence for a given bin (darker shades signify under-confidence, while brighter red colors denote over-confidence).

\begin{table}[ht]
    \caption{Calibration error evaluated using three estimators for each of the classes within CIFAR-10. The values are averaged over four seeds. The numbers in the brackets represent the ranking order. }
    \centering
    \vskip 0.1in
    \resizebox{\textwidth}{!}{
    \begin{tabular}{cccccccccccc}
        \toprule
        Model & Metric & Class 0 & Class 1 & Class 2 & Class 3 & Class 4 & Class 5 & Class 6 & Class 7 & Class 8 & Class 9 \\
        \midrule
        \multirow{3}{*}{PreResNet56} 
        & $\widehat{\operatorname{CE}}_{\mathrm{KL}} \times 100 $   & 1.58 (6) & 0.91 (1) & 2.13 (8) & 4.28 (10) & 1.78 (7) & 3.59 (9) & 1.30 (4) & 1.09 (2) & 1.12 (3) & 1.34 (5) \\
        & $\widehat{\operatorname{CE}_{2}^{2}} \times 1000 $   & 7.25 (6) & 4.04 (1) & 9.27 (8) & 16.96 (10) & 8.26 (7) & 14.58 (9) & 5.34 (3) & 5.77 (4) & 5.31 (2) & 5.99 (5) \\
        & $\operatorname{CE}_1 \times 1000  $  & 6.25 (6) & 3.36 (1) & 7.65 (8) & 15.62 (10) & 6.51 (7) & 12.76 (9) & 4.34 (3) & 3.96 (2) & 4.52 (4) & 4.97 (5) \\
        \midrule
        \multirow{3}{*}{WideResNet28x10} 
        & $\widehat{\operatorname{CE}}_{\mathrm{KL}} \times 100 $  & 0.86 (5) & 0.72 (3) & 1.43 (8) & 2.76 (10) & 0.96 (7) & 2.43 (9) & 0.71 (2) & 0.53 (1) & 0.75 (4) & 0.93 (6) \\
        & $\widehat{\operatorname{CE}_{2}^{2}} \times 1000 $   & 4.80 (6) & 3.65 (2) & 7.52 (8) & 13.39 (10) & 5.22 (7) & 11.76 (9) & 3.73 (3) & 2.91 (1) & 4.01 (4) & 4.73 (5) \\
        & $\operatorname{CE}_1 \times 1000 $   & 3.47 (5) & 2.80 (3) & 5.42 (8) & 11.61 (10) & 3.57 (7) & 10.12 (9) & 2.50 (2) & 1.83 (1) & 2.91 (4) & 3.49 (6) \\
        \bottomrule
    \end{tabular}
    }
    \label{tab:classwise_metrics}
\end{table}   

\end{document}